\documentclass[journal]{IEEEtran}
\newcommand{\figref}[1]{{Fig.~\ref{#1}}}
\newcommand{\tabref}[1]{{Table.~\ref{#1}}}

\newcommand{\secref}[1]{Section.~\ref{#1}}

\usepackage{caption}
\newcommand{\figcaption}[1]{\def\@captype{figure}\caption{#1}}
\newcommand{\tblcaption}[1]{\def\@captype{table}\caption{#1}}
\usepackage[final]{graphicx}
\usepackage{here}
\usepackage{color}

\usepackage{bm}
\usepackage{amsmath}
\usepackage{relsize}
\usepackage[subrefformat=parens]{subcaption}
\usepackage{siunitx}
\usepackage{multirow}

\usepackage{placeins}

\usepackage{ifpdf}
\usepackage{cite}
\usepackage{algorithmic}

\usepackage{array}

\usepackage{url}
\begin{document}

%
% paper title
% Titles are generally capitalized except for words such as a, an, and, as,
% at, but, by, for, in, nor, of, on, or, the, to and up, which are usually
% not capitalized unless they are the first or last word of the title.
% Linebreaks \\ can be used within to get better formatting as desired.
% Do not put math or special symbols in the title.
\title{BEATLE---Self-Reconfigurable Aerial Robot: Design, Control and Experimental Validation}
%
%
% author names and IEEE memberships
% note positions of commas and nonbreaking spaces ( ~ ) LaTeX will not break
% a structure at a ~ so this keeps an author's name from being broken across
% two lines.
% use \thanks{} to gain access to the first footnote area
% a separate \thanks must be used for each paragraph as LaTeX2e's \thanks
% was not built to handle multiple paragraphs
%

\author{Junichiro\;Sugihara,
        Moju\;Zhao,
        Takuzumi\;Nishio,
        Kei\;Okada,
        and\;Masayuki\;Inaba% <-this % stops a space
\thanks{Junichiro\;Sugihara\;(corresponding\;author),\;Takuzumi\;Nishio,\;Kei\;Okada,\;and Masayuki\;Inaba\;are\;with\;the\;Department\;of\;Mechano-Infomatics,\;The\;University\;of\;Tokyo,\;Bunkyo-ku,\;Tokyo\;113-8656,\;Japan\;(e-mail:j-sugihara@jsk.imi.i.u-tokyo.ac.jp\;nishio@jsk.imi.i.u-tokyo.ac.jp\;k-okada@jsk.imi.i.u-tokyo.ac.jp\;inaba@jsk.imi.i.u-tokyo.ac.jp).}
% <-this % stops a space

\thanks{Moju\;Zhao\;is\;with\;the\;Department\;of\;Mechanical-Engineering,\;The\;University\;of\;Tokyo,\;Bunkyo-ku,\;Tokyo\;113-8656,\;Japan\;(e-mail:chou@jsk.imi.i.u-tokyo.ac.jp).}% <-this %
        % stops a space
% \thanks{Manuscript received April 19, 2005; revised August 26, 2015.}
}
% note the % following the last \IEEEmembership and also \thanks - 
% these prevent an unwanted space from occurring between the last author name
% and the end of the author line. i.e., if you had this:
%
% \author{....lastname \thanks{...} \thanks{...} }
%                     ^------------^------------^----Do not want these spaces!
%
% a space would be appended to the last name and could cause every name on that
% line to be shifted left slightly. This is one of those "LaTeX things". For
% instance, "\textbf{A} \textbf{B}" will typeset as "A B" not "AB". To get
% "AB" then you have to do: "\textbf{A}\textbf{B}"
% \thanks is no different in this regard, so shield the last } of each \thanks
% that ends a line with a % and do not let a space in before the next \thanks.
% Spaces after \IEEEmembership other than the last one are OK (and needed) as
% you are supposed to have spaces between the names. For what it is worth,
% this is a minor point as most people would not even notice if the said evil
% space somehow managed to creep in.

% The paper headers
\markboth{Journal of \LaTeX\ Class Files,~Vol.~14, No.~8, August~2015}%
{Shell \MakeLowercase{\textit{et al.}}: Bare Demo of IEEEtran.cls for IEEE Journals}
% The only time the second header will appear is for the odd numbered pages
% after the title page when using the twoside option.
%
% *** Note that you probably will NOT want to include the author's ***
% *** name in the headers of peer review papers.                   ***
% You can use \ifCLASSOPTIONpeerreview for conditional compilation here if
% you desire.

% If you want to put a publisher's ID mark on the page you can do it like
% this:
%\IEEEpubid{0000--0000/00\$00.00~\copyright~2015 IEEE}
% Remember, if you use this you must call \IEEEpubidadjcol in the second
% column for its text to clear the IEEEpubid mark.

% use for special paper notices
%\IEEEspecialpapernotice{(Invited Paper)}

% make the title area
\maketitle

% As a general rule, do not put math, special symbols or citations
% in the abstract or keywords.
\begin{abstract}
Modular self-reconfigurable robots (MSRRs) offer enhanced task flexibility by constructing various structures suitable for each task. However, conventional terrestrial MSRRs equipped with wheels face critical challenges, including limitations in the size of constructible structures and system robustness due to elevated wrench loads applied to each module. In this work, we introduce a Aerial MSRR (A-MSRR) system named BEATLE, capable of merging and separating in-flight. BEATLE can merge without applying wrench loads to adjacent modules, thereby expanding the scalability and robustness of conventional terrestrial MSRRs. In this article, we propose a system configuration for BEATLE, including mechanical design, a control framework for multi-connected flight, and a motion planner for reconfiguration motion. The design of a docking mechanism and housing structure aims to balance the durability of the constructed structure with ease of separation. Furthermore, the proposed flight control framework achieves stable multi-connected flight based on contact wrench control. Moreover, the proposed motion planner based on a finite state machine (FSM) achieves precise and robust reconfiguration motion. We also introduce the actual implementation of the prototype and validate the robustness and scalability of the proposed system design through experiments and simulation studies.
\end{abstract}

% Note that keywords are not normally used for peerreview papers.

% For peer review papers, you can put extra information on the cover
% page as needed:
% \ifCLASSOPTIONpeerreview
% \begin{center} \bfseries EDICS Category: 3-BBND \end{center}
% \fi
%
% For peerreview papers, this IEEEtran command inserts a page break and
% creates the second title. It will be ignored for other modes.
\IEEEpeerreviewmaketitle

% Introduction
\section{Introduction} \label{sec:introduction}
\begin{figure}[!t]
 \begin{center}
   \includegraphics[width=\columnwidth]{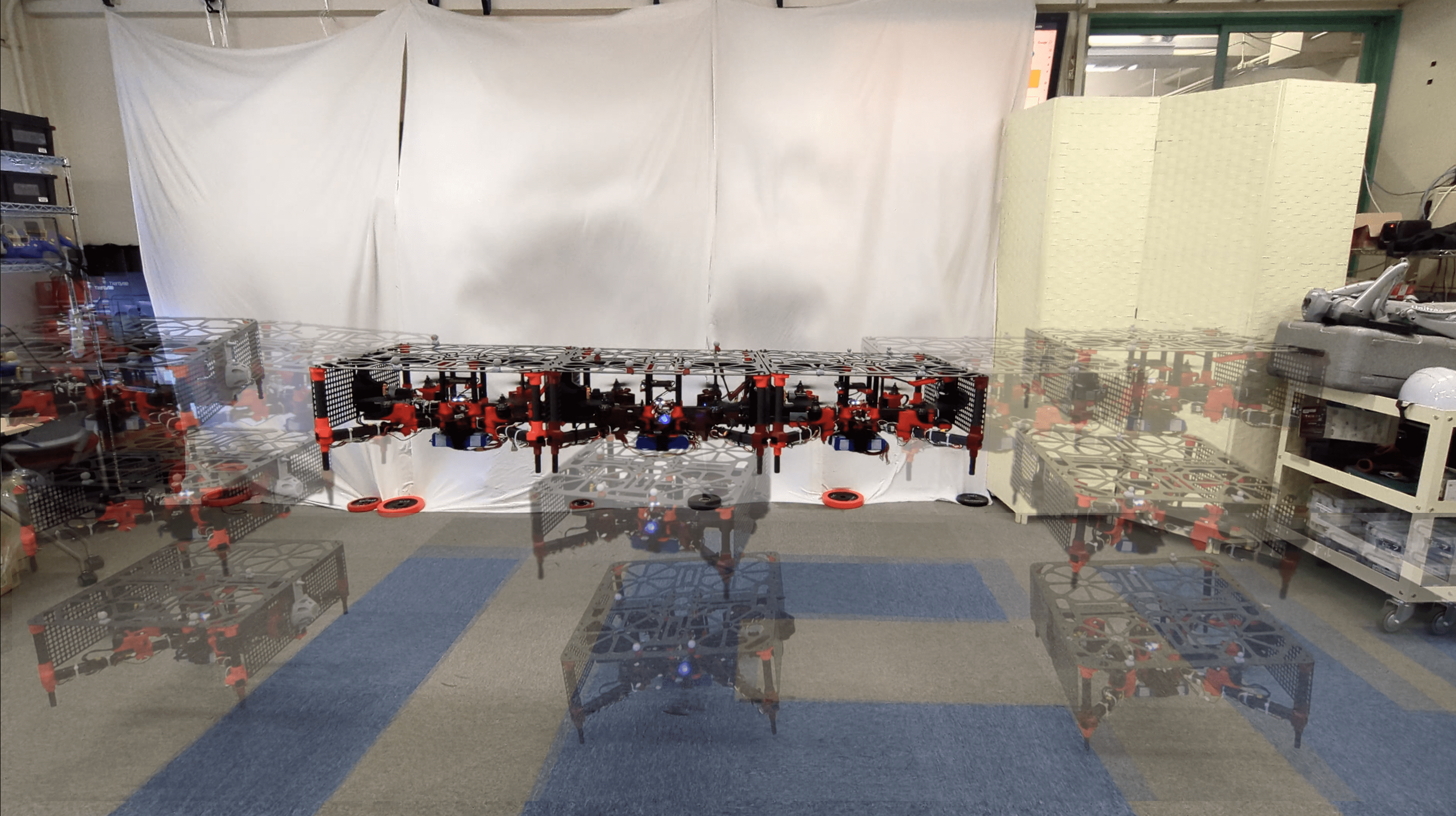}
 \end{center}
 \caption{
The proposed robot platform BEATLE: \textbf{BE}nding-moment-free self-reconfigurable \textbf{A}erial robo\textbf{T} modu\textbf{LE}. This chronophotography visualizes reconfiguration motion of BEATLE.}
   \label{figure:BEATLE_main}
\end{figure}
\IEEEPARstart{I}{n recent} years, significant progress has been made in the field of modular self-configurable robots (MSRRs)\cite{Jorgensen2004,Yim2007,Zykov2007,Chennareddy2017,Brunete2017}, which possess the ability to autonomously merge to new create structures. The principal objective of MSRRs is to address various tasks utilizing a single type of robot, achieved through the construction of structures tailored to each task. For example, mobility can be enhanced by constructing wheel or leg structures\cite{Wei2011,Davey2012,Liu2023}, while manipulation capabilities can be augmented through the creation of arm structures\cite{Liang2020,Swissler2020,Yuxiao2022}. Furthermore, the integration of multiple bodies can yield the formation of bridge-like structures\cite{Tosun2018,Yuxiao2022}, thereby facilitating the movement of other robots.\par
However, these terrestrial MSRRs encounter the drawback of being constrained in the scale of structures they can construct. This limitation is particularly evident in cantilever configurations such as arm and bridge structures. In the scenario of a cantilever arrangement, a module imposes a contact wrench on adjacent modules due to its own weight. Therefore, considering the module on the edge as the $1st$ module, the $i-th$ module is subject to a wrench load induced by the weight of $(i-1)$ bodies. Consequently, as the scale of the intended structure increases, the impact of contact wrench becomes more pronounced. These contact wrenches can lead to overloading of actuators and mechanical components, thereby reducing system responsiveness and structural robustness. Hence, to address this problem, it is necessary for each module to compensate for its own weight and control its orientation independently.\par

In this study, we introduce an aerial Modular Self-Reconfigurable Robot (A-MSRR) platform named BEATLE\figref{figure:BEATLE_main}, designed for autonomous merging and separation in mid-air. Similar to terrestrial MSRRs, BEATLE enhances available force and torque by rigidly interconnecting with one another, thereby augmenting task capabilities. Furthermore, the interlinking of aerial modules enables the autonomous assembly of structures such as bridges and ladders at elevated altitudes. Although terrestrial MSRRs are equipped with rotational joints\cite{Chennareddy2017} at the docking points, aerial robots can execute diverse tasks without joints\cite{Watson2022,Martinez2023} since their bodies possess inherent three-dimensional mobility. Additionally, BEATLE possesses the capability to exert force in an arbitrary direction through its thrust vectoring rotor, enabling it to compensate its own weight in any orientation without imposing torque loads on neighboring bodies, during a multi-connected state. Hence, BEATLE facilitates amalgamation with an arbitrary number of moudules, substantially enhancing the scalability of MSRR.

\subsection{Related works}
\subsubsection{Docking mechanism}
Docking mechanism is one of the most crucial hardware components in MSRRs, and various coupling methods have been proposed in prior research. For example, the A-MSRR presented in \cite{Saldana2018} features a magnetic docking mechanism. Magnetic coupling proves suitable for A-MSRRs as magnetic force can compensate for misalignment between two docking parts, simplyfing aerial docking. However, \cite{Saldana2018} lacks the capability for undocking motion due to the absence of a mechanism to release magnetic connections. Although the A-MSRR in \cite{moddessemble}, equipped with a much smaller magnetic connector, can perform aerial undocking motion by applying a momentary bending moment to the docking part with rotor thrusts, this smaller connector conversely complicates the docking motion. In contrast, TRADY, an A-MSRR proposed in our previous work \cite{sugihara2023}, achieves both docking and undocking motion by employing a pole-switching mechanism with a servo motor.\par
However, a remaining issue with magnetic connections is the lack of resistance against bending moments. Particularly, to enable high-torque activities and the assembly of large structures, a more robust mechanical coupling is necessary.
Regarding mechanical connections, certain mechanisms for terrestrial MSRRs adopt this approach. For instance, in \cite{Behnam2006}, a docking mechanism comprised of locking ``jaws'' capable of hooking onto the corresponding ``jaws'' on the other side is proposed. Additionally, a mechanism consisting of movable pins that can be inserted into the holes of other modules is also suggested in \cite{Wolfe2012}. However, both these ``jaws'' and pins are relatively small compared to the overall body size, which may limit their practical durability.
Therefore, this article introduces a hybrid docking mechanism composed of both magnetic and mechanical connectors. The mechanical connection is achieved by inserting sticks which are supported by durable structure. Additionally, to improve the rigidity of the assembled structure and ensure safer reconfiguration motion, a housing design for aerial modules is proposed.

\subsubsection{Control method}
\begin{figure}[!t]
 \begin{center}
   \includegraphics[width=\columnwidth]{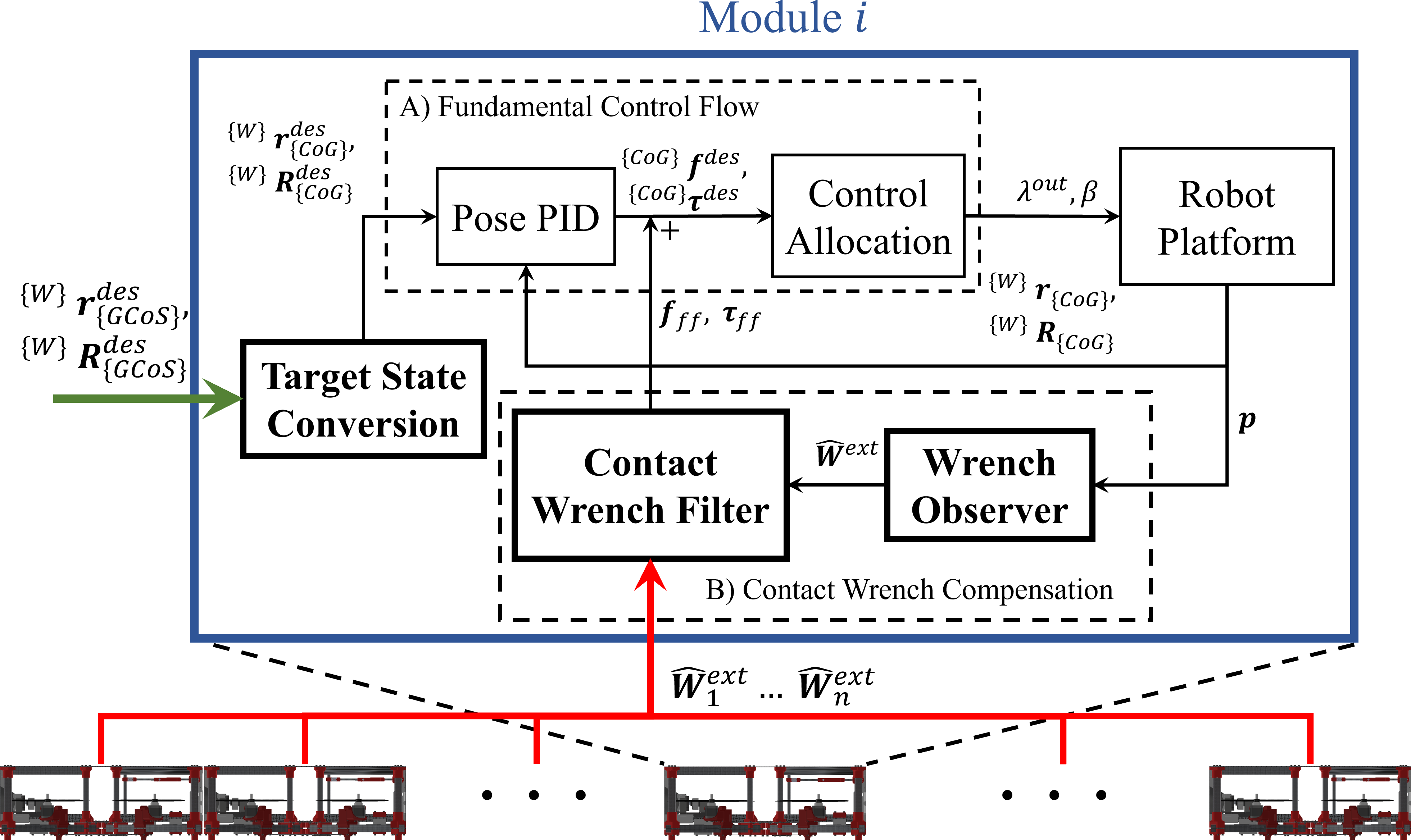}
 \end{center}
 \caption{The proposed control framework for BEATLE: (A) Fundamental Control Flow and (B) Contact Wrench Compensation. Blocks highlighted in bold font denote processes essential for multi-connected flight. Green arrows signify reference inputs from upstream sources, while red arrows indicate information exchange with other modules. Each blue block represents an individual module.}
   \label{fig:framework}
\end{figure}
\begin{figure*}[!t]
 \begin{center}
   \includegraphics[width=1.8\columnwidth]{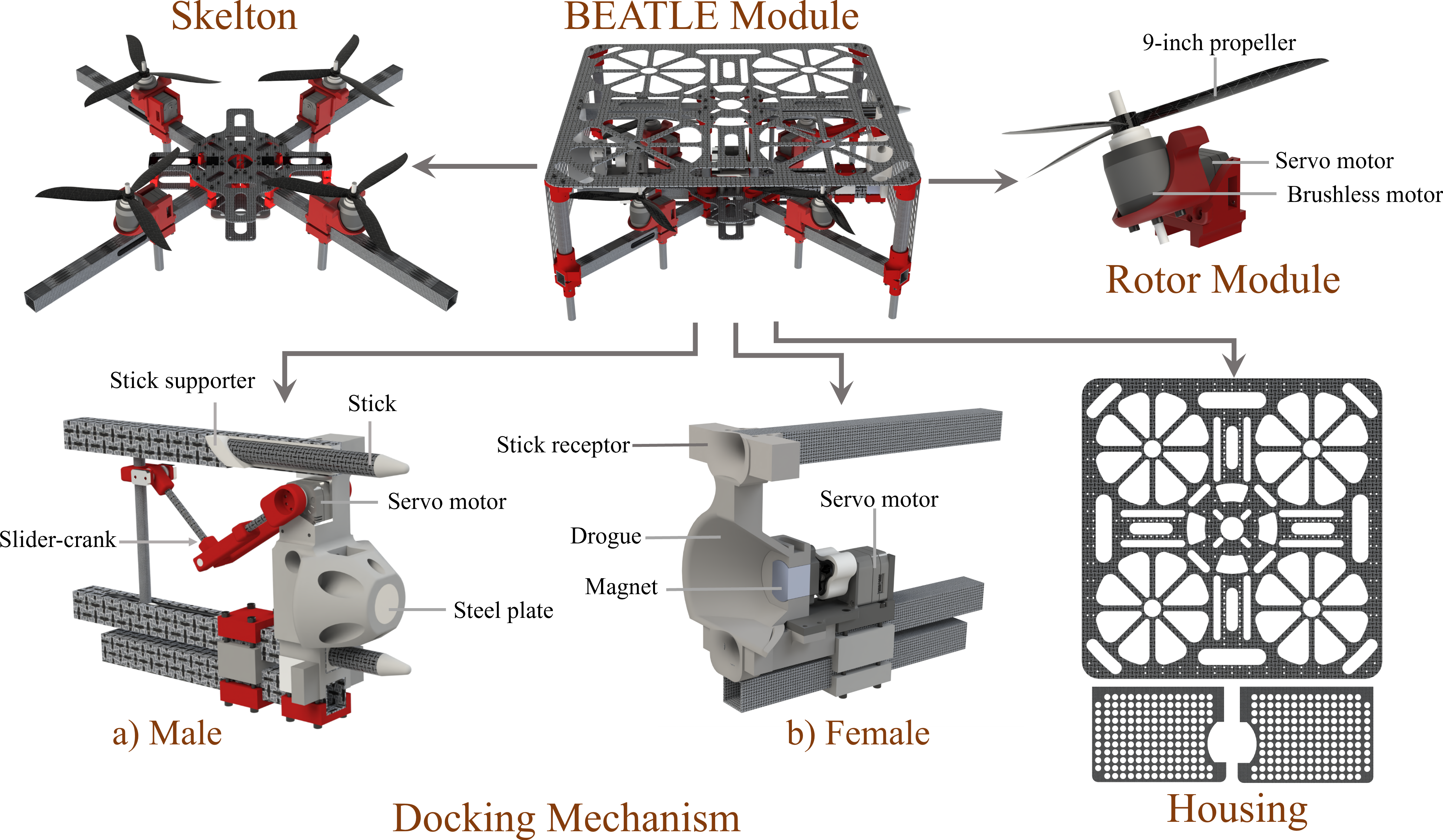}
 \end{center}
 \caption{The overview of the proposed design for BEATLE. Top: visualization of BEATLE module and its fundamental components. Bottom: visualization of docking mechanisms and housing plates where (a) represents the male side and (b) represents the female side.}
   \label{figure:full_design}
\end{figure*}
The key points regarding the stability and scalability of the A-MSRR system include robustness during multi-connected flight and stability of aerial reconfiguration motion. These attributes are characterized by the controllability of each module. These attributes depend on the control degree of freedom (CDoF) of the assembled structure and each module.
Concerning the controllability of the assembled structure, A-MSRRs proposed in \cite{Saldana2018, moddessemble, Flightarray1, Flightarray2},are underactuated systems during both single flight and multi-connected flight. According to the general control theory for underactuated aerial robot \cite{Kumar2012}, translational control is achieved through the rotational motion of the airframe. However, as the scale of the assembled structure increases, the responsiveness of rotational motion decreases. Hence, when more than a certain number of modules are connected, flight control becomes critically unstable. On the other hand, in the case of TRADY \cite{sugihara2023}, while each module is underactuated, the merged structure is over-actuated. Consequently, TRADY can address the instability caused by the low responsiveness of rotational motion.\par

On the other hand, concerning the controllability of each module, all previously proposed A-MSRRs are underactuated. These underactuated A-MSRRs cannot explicitly control the inter-module contact forces that arise during multi-connected flight because underactuated modules do not have extra CDoF. Therefore, the flight control of A-MSRR in the merged state is achieved by employing a dynamics model that considers the entire connected structure as a single aircraft. However, a significant issue with this approach is that the dynamics model required for control needs updating during reconfiguration. This updating results in discontinuous changes in the target thrust during reconfiguration. These discrete changes lead to instability in the aircraft, such as sudden ascents or descents \cite{Saldana2018,sugihara2023}.\par

Another challenge arises from the accumulation of unexpected contact wrench between adjacent modules. This is because the control process is decentralized, with each module independently carrying out its control procedures. Consequently, variations in target control outputs among modules arise from deviations in self-state estimation and subtle disparities in aircraft characteristics. The interference between modules resulting from this discrepancy in target outputs leads to the accumulation of contact wrenches. As the number of modules increases, these contact wrenches become more pronounced, destabilizing flight control and reconfiguration operations, thus significantly impacting the scalability of the system.\par

Therefore, BEATLE proposed in this study features a thrust configuration, enabling each module to perform overactuated control independently. This aircraft configuration, along with the proposed control framework, enables explicit control of inter-module contact wrenches during multi-connected flight. Thus, the problems of dynamics model switching and contact wrench accumulation that were problems with previous A-MSRRs can be solved, greatly improving system stability and scalability. This control framework is summarized in \figref{fig:framework}.
\subsection{Contributions of this work}
In summary, this paper presents an extended A-MSRR system, offering the following novel contributions:
\vspace{-2pt}
 \begin{enumerate}
  \item We propose a mechanical design for A-MSRR aimed at constructing structures with heightened rigidity.
  \item We introduce a flight control framework facilitating scalable multi-connected flight and ensuring stable self-reconfiguration based on contact wrench control.
  \item We validate the efficacy of the proposed A-MSRR system through a comprehensive evaluation encompassing real-machine experiments and simulation studies.
 \end{enumerate}
 \vspace{-2pt}
To the best of our knowledge, this is the first work that introduces an A-MSRR capable of conducting multi-connected flight utilizing a contact wrench controller.
\subsection{Notation}
From this section, nonbold lowercase symbols (e.g., $m$) represent scalars, and bold symbols (e.g., $\bm{u}$) represent vectors. Superscripts (e.g, ${}^{\lbrace{CoG}\rbrace}\bm{p}$) represent the frame in which the vector or matrics is expressed, and subscripts represent the target frame or an axis, e.g.,${}^{\lbrace W\rbrace}\bm{r}_{\lbrace CoG \rbrace}$ represents a vector point from ${\lbrace W\rbrace}$ to ${\lbrace CoG\rbrace}$ w.r.t. ${\lbrace W\rbrace}$, whereas $u_{x}$ denotes the $x$ component of the vector $\bm{u}$. Furthermore, superscripts (e.g., $^{t}\bm{f}$, $\bm{\Lambda}^{T}$) represent transposed vectors.
\subsection{Organization}
\secref{sec:design} introduces the module configuration method, covering thrust configuration and mechanical design. Subsequently, \secref{sec:control} introduces the design of basic flight controllers for the modules. Following this, \secref{sec:planning} describes the motion planning method for reconfiguration operations, while \secref{sec:multi} proposes a method to convert target state and a contact wrench control framework for multi-connected flight. Then, \secref{sec:simulation} evaluates the proposed methods through simulations, and \secref{sec:experiment} elaborates on the mechanical implementation of the actual system and the experimental results obtained using it  before concluding in \secref{sec:conclusion}.

%Design
\section{Design}\label{sec:design}
\subsection{Rotor Configuration}
As highlighted in \secref{sec:introduction}, achieving overactuation for BEATLE necessitates the use of vectoring rotors. While the minimum requirement is three rotors, this study opts for four rotors to enhance redundancy and maintain symmetry across the entire module structure. As depicted in \figref{figure:full_design}, each rotor can rotate its thrust direction about an axis parallel to its rotor arm within the range of $[-\frac{\pi}{2}, \frac{\pi}{2}]$. To simplify the rotor structure, we employ the direct drive method for rotor tilting, foregoing the use of gears or pulleys in the transmission mechanism. In summary, with four rotors and four servos, the BEATLE module has a total control input dimension of eight.
\subsection{Docking Mechanism}
\begin{figure}[!t]
 \begin{center}
   \includegraphics[width=0.9\columnwidth]{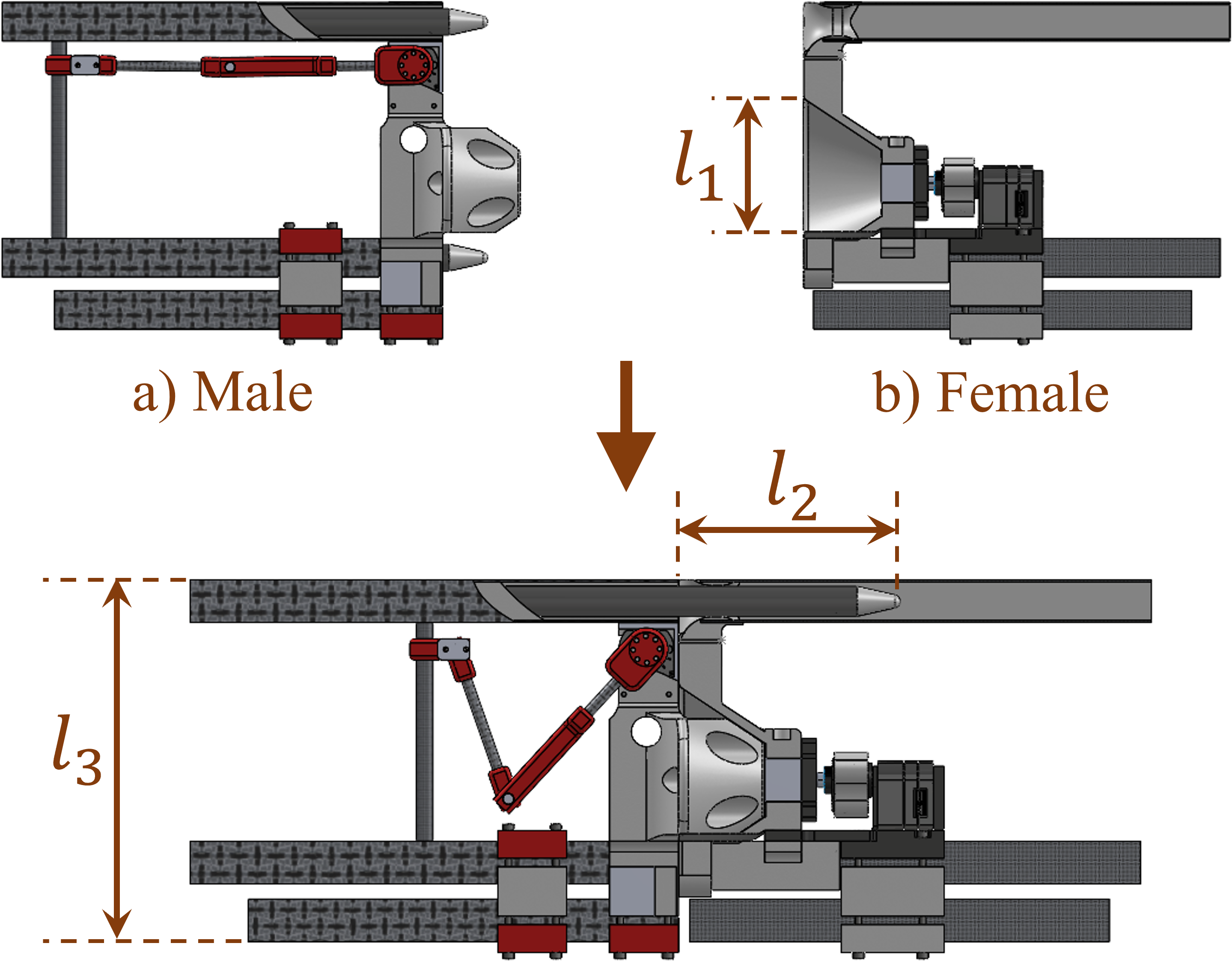}
 \end{center}
 \caption{The diagram illustrating the operation of the coupling mechanism. Some parts are depicted in cross-sectional view for ease of viewing. $l_{1}$-$l_{3}$ are key parameters of design.}
   \label{fig:docking}
\end{figure}
The entire coupling system design is illustrated in \figref{figure:full_design}, with the male side depicted in a) and the female side in b). Furthermore, the operation of the docking system is illustrated in \figref{fig:docking}. This system features two key aspects.\par
Initially, the female side mechanism is equipped with a switchable permanent magnet module as proposed in \cite{sugihara2023}, which is driven by a servo motor. The tensile strength of this magnetic connection is \SI{220}{N}. Additionally, the female side mechanism features a cone-shaped structure known as a drogue, which compensates for alignment errors between modules. $l_{1}$ in \figref{fig:docking} corresponds to the tolerable range of error that can be compensated by the drogue. In this study, $l_{1}$ is set to \SI{70}{mm} based on the position control accuracy obtained through experiments.\par

Secondly, the male side incorporates a stick insertion system utilizing a slider-crank mechanism, which is also driven by a servo motor. These sticks are composed of CFRP pipe and PLA core, and they are supported by sockets firmly fixed to BEATLE's skeleton. This structure efficiently converts most of the applied bending load on the sticks into tensile force. Thus, the torque resistance of the proposed mechanism primarily depends on the tensile strength of the sticks. In the case of CFRP, its tensile strength is \SI{2200}{MPa} and is durable enough for the expected load. $l_{2}$ and $l_{3}$ are parameters representing the length of the inserted stick and the height of the coupling mechanism, respectively, with $l_{2}$ dependent on $l_{3}$. In this study, $l_{3}$ is set to \SI{170}{mm} as the minimum value to ensure clearance for the propeller, and consequently, $l_{2}$ was set to \SI{110}{mm}.

\subsection{Housing}
BEATLE's housing shown in \figref{figure:full_design} serves three main purposes: firstly, to enhance the rigidity of the docking connection by expanding the contact area; secondly, to prevent propeller collision during reconfiguration motion; and lastly, to increase the utility of structures constructed with BEATLE, such as bridges. However, as the housing surrounds the rotor, it potentially obstructs air intake and affects thrust efficiency. To address this issue, appropriate lightening holes are made in the housing to ensure airflow. In this study, approximately \SI{65}{\%} of the housing's footprint is cut out. \tabref{table:thrust_housing} compares the thrust during hovering with and without the rotor covered by the housing. Since the housing weighs \SI{0.4}{kg}, the thrust loss of each rotor caused by aerodynamic interference is calculated as \SI{0.7}{N}. This value is \SI{6.2}{\%} of the hovering thrust, which is not considered to be a significant problem.
\begin{table}[b]
  \renewcommand{\arraystretch}{1.3}
  \caption{Effect of housing on thrust.}
  \centering
 \begin{tabular}{|c|c|c|}
  \hline
  &With housing&Without housing \\
  \hline
  Hovering Thrust & \SI{10.5}{N} & \SI{12.2}{N}\\ \hline
 \end{tabular}
 \label{table:thrust_housing}
\end{table}
\renewcommand{\arraystretch}{1.0}

%Control
\section{Fundamental Flight Control}\label{sec:control}
In this section, we present the design of the fundamental flight controller for each module. The process introduced in this section corresponds to ``Fundamental Control Flow'' in \figref{fig:framework}.
\subsection{Single Module Dynamics Model}
\begin{figure}[!t]
 \begin{center}
   \includegraphics[width=0.8\columnwidth]{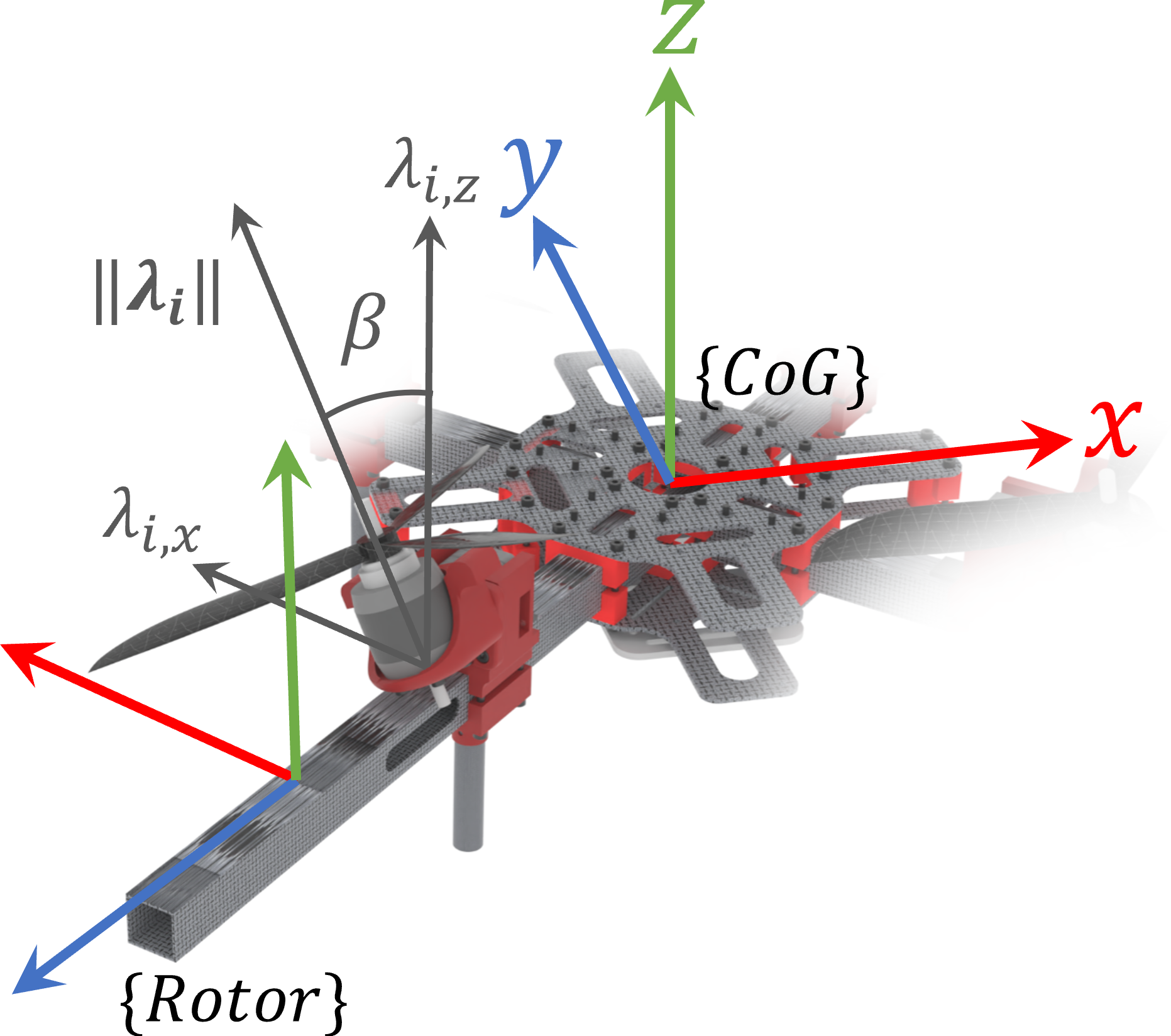}
 \end{center}
 \caption{Coordinate model for BEATLE module.}
   \label{fig:beatle_kinematics}
\end{figure}

\begin{table}[!t]
  \renewcommand{\arraystretch}{1.3}
  \caption{Definition of quantities for dynamics model.}
  \centering
 \begin{tabular}{|c|c|}
  \hline
  Symbol&Definition \\
  \hline
  $m$& Mass of body\\
  $\text{\textbf{I}}$& Moment of inertia of body\\
  $\bm{r}$& Position of CoG\\
  $\bm{\alpha} =\lbrack\begin{array}{ccc}
                  \theta&\phi &\psi \\
                       \end{array}\rbrack$ & Euler angles (roll, pitch, yaw)\\
  $\bm{p_{i}}$& Positiopn of i-th rotor\\
  $\sigma_{i} \in \left[-1, \;1\right]$ & Counter torque of each rotor\\
  $c$ & Yaw-thrust ratio\\
  $\bm{f}$& Resultant force\\
  $\bm{\tau}$& Resultant torque\\
  $\bm{\omega}$&Angular velociy\\
  \hline
 \end{tabular}
 \label{table:dynamics_param}
\end{table}
\renewcommand{\arraystretch}{1.0}
The coordinates model for BEATLE is depicted in \figref{fig:beatle_kinematics}, with each quantity defined as shown in \tabref{table:dynamics_param}. \par 
Originally, the dynamics model including rotor thrust vectoring are nonlinear system. However, \cite{Kamel2018} proposed a method to linearize the dynamics model by assuming that rotor thrust vectoring occurs momentarily and the intermediate rotational motion can be disregarded. In this study, we also adopt this approximation. \par
First of all, as illustrated in \figref{fig:beatle_kinematics}, the rotor thrust of the $i-th$ rotor can be expressed as following:
\begin{equation}\label{equation:lambda_i}
 {}^{\lbrace Rotor_{i} \rbrace}\bm{\Lambda}_{i} = \left[\begin{array}{ccc}
                     \lambda_{i,x} & 0 & \lambda_{i,z}^{}
                \end{array}
\right]^T,
\end{equation}
where $\bm{\Lambda}_{i}$ is the thrust force vector of $i-th$ rotor, while $\lambda_{i,x}$ and $\lambda_{i,z}$ are rotor thrust elements aligned with the $x$ and $z$ axis of $\lbrace Rotor_{i} \rbrace$, respectively.\par
% Then, the $i-th$ thrust vector $\bm{\Lambda_{i}}$ in $\lbrace Rotor_{i} \rbrace$ can be expressed as follows:
% \begin{equation}\label{equation:rotor_arm_force}
%  {}^{\lbrace {}^{i}Rotpor \rbrace}\bm{\Lambda}_{i} = 
% \left(\begin{array}{c}
%   0\\
%    \lambda_{i,x}\\
%    \lambda_{i,z}
%            \end{array}
% \right)
%  = B\bm{\lambda_{i}},
% \end{equation}
% where
% \begin{equation}\label{equation:B}
%  B = \left(\begin{array}{cc}
%   0 & 0\\
%   1 & 0 \\
%    0&1\\
%            \end{array}
% \right).
% \end{equation}
% Note that the $B$ in \eqref{equation:B} is a mapping from the thrust set $\lambda$ to the three-dimensional thrust vector $\Lambda$.\par
Here, the coordinates conversion of thrust vector from $i-th$ rotor arm coordinates $\lbrace Rotor_{i} \rbrace$ to center-of-gravity coordinates $\lbrace CoG \rbrace$ can be written as follows:
\begin{align} \label{equation:integrated_r}
{}^{\lbrace CoG \rbrace}\bm{\Lambda}_{i}& = {}^{\lbrace CoG \rbrace}R_{\lbrace Rotor_{i} \rbrace} {}^{\lbrace Rotor_{i} \rbrace}\bm{\Lambda}_{i} % \notag\\ 
%  & = {}^{\lbrace CoG \rbrace}R_{\lbrace Rotor_{i} \rbrace} B \bm{\lambda_{i}}\notag\\
% & = R'_{i}\bm{\lambda_{i}}
\end{align}
where ${}^{\lbrace CoG \rbrace}R_{\lbrace Rotor_{i} \rbrace}$ is the transformation matrix from $\lbrace Rotor_{i} \rbrace$ to $\lbrace CoG \rbrace$.\par
Based on \eqref{equation:integrated_r}, the wrench-force
${}^{\lbrace{CoG}\rbrace}\bm{f}$ and torque
${}^{\lbrace{CoG}\rbrace}\bm{\tau}$ )  can be written as

\begin{equation}\label{equation:force}
 {}^{\lbrace{CoG}\rbrace}\bm{f} = \sum\limits_{i=1}^{4}{}^{\lbrace CoG \rbrace}\bm{\Lambda}_{i},
\end{equation}
\begin{equation}\label{equation:torque}
 {}^{\lbrace{CoG}\rbrace}\bm{\tau} = \sum\limits_{i=1}^{4}\left(\left[{}^{\lbrace{CoG}\rbrace}\bm{p}_{i}\times\right] -c\:\sigma_{i}{}^{\lbrace{CoG}\rbrace}\right){}^{\lbrace CoG \rbrace}\bm{\Lambda}_{i}.
\end{equation}
From \eqref{equation:force} and \eqref{equation:torque}, the
translational and rotational dynamics of a multirotor unit are given by the Newton-Euler
equation as followings:
\begin{equation}\label{equation:translation}
 m{}^{\lbrace W\rbrace}\ddot{\bm{r}}_{\lbrace CoG \rbrace} =
 {}^{\lbrace W\rbrace}R_{\lbrace CoG \rbrace}{}^{\lbrace{CoG}\rbrace}\bm{f}-m\bm{g},
\end{equation}
\begin{equation}\label{equation:rotation}
\begin{split}
 {}^{\lbrace{CoG}\rbrace}I{}^{\lbrace{CoG}\rbrace}\dot{\bm{\omega}}
  &=
 {}^{\lbrace{CoG}\rbrace}\bm{\tau}+\bm{\sigma}{}^{\lbrace{CoG}\rbrace}\bm{f}\\
 &\quad -
 {}^{\lbrace{CoG}\rbrace}\bm{\omega}\times{}^{\lbrace{CoG}\rbrace}I{}^{\lbrace{CoG}\rbrace}\bm{\omega},
\end{split}
\end{equation}
where ${\lbrace W\rbrace}$ frame represent the world coordinate system and $\bm{g}$ is gravity acceleration vector.
\subsection{Pose PID Control}
Considering equations \eqref{equation:force} and \eqref{equation:torque}, a common PID controller is applied to obtain the desired wrench on the $\lbrace CoG \rbrace$ as follows:
\begin{equation}\label{equation:des_f}
\begin{split}
  \bm{f}^{des} = m\;{}^{\lbrace W\rbrace}&\bm{R}_{\lbrace CoG \rbrace}^{-1}\left(K_{f, p}\bm{e_{r}} +K_{f,i}\int\bm{e_{r}}dt
                       +K_{f, d}\dot{\bm{e_{r}}}\right)\\ &+ \bm{f}^{ff},
\end{split}
\end{equation}
\begin{equation}\label{equation:des_t}
 \begin{split}
 \bm{\tau}^{des} =\;{}^{\lbrace{CoG}\rbrace}\bm{I}\left(K_{\tau, p}\bm{e_{\alpha}} +K_{\tau, i}\int\bm{e_{\alpha}}dt
  +\;K_{\tau, d}\dot{\bm{e_{\alpha}}}\right) \\
   +
  {}^{\lbrace{CoG}\rbrace}\bm{\omega}\times\;{}^{\lbrace{CoG}\rbrace}\bm{I}\;{}^{\lbrace{CoG}\rbrace}\bm{\omega}  + \bm{\tau}^{ff},
   \end{split}
\end{equation}
where $\bm{e_{r}}$ is a position error defined as $\bm{e_{r}} =
{}^{\lbrace W \rbrace}\bm{r}^{des}_{\lbrace CoG \rbrace} - {}^{\lbrace W \rbrace}\bm{r}_{\lbrace CoG \rbrace}$, and
$\bm{K}_{. , p}$, $\bm{K}_{. , i}$, $\bm{K}_{. , d}$ are gains for
controller. Additionally, $f^{ff}$ and $\tau^{ff}$ represent the feedforward terms generated in the contact wrench compensation flow, as depicted in \figref{fig:framework} B).

\subsection{Control Allocation}
Using \eqref{equation:force} and
\eqref{equation:torque}, allocation from the thrust force $\bm{\lambda}$
to the resultant wrench ${}^{\lbrace{CoG}\rbrace}\bm{W}$ can be given by following:
\begin{equation}\label{equation:allocation}
{}^{\lbrace{CoG}\rbrace}\bm{W} = 
 \left(\begin{array}{c}
  {}^{\lbrace{CoG}\rbrace}\bm{f} \\
        {}^{\lbrace{CoG}\rbrace}\bm{\tau}
       \end{array}
 \right)
 = Q{}^{\lbrace CoG \rbrace}\bm{\Lambda}_{i},
\end{equation}
where
\begin{equation}
 Q =    \left(\begin{array}{c}
  E\\
  \left[{}^{\lbrace{CoG}\rbrace}\bm{p}_{i}\times\right]-c\:\sigma_{i}{}^{\lbrace{CoG}\rbrace}
       \end{array}
        \right),
\end{equation}
\begin{equation}
 \bm{\Lambda} = \left[
\begin{array}{cccc}
 {}^{\lbrace CoG \rbrace}\bm{\Lambda}_{1}^T& {}^{\lbrace CoG \rbrace}\bm{\Lambda}_{2}^T& {}^{\lbrace CoG \rbrace}\bm{\Lambda}_{3}^T &{}^{\lbrace CoG \rbrace}\bm{\Lambda}_{4}^T \\
\end{array}
\right]^{T}.
\end{equation}
% Next, using \eqref{equation:integrated_r}, \eqref{equation:allocation} can be rearranged as follows:
% \begin{align} \label{equation:wrench_allocation}
% {}^{\lbrace{CoG}\rbrace}\bm{W}
%  &= 
% Q
% \left[
% \begin{array}{ccc}
%  \left( R'_{1} \bm{\lambda}_{1} \right)^{T}& \dots & \left( R'_{4} \bm{\lambda}_{4}\right)^{T}
% \end{array}
% \right]^{T}\notag\\
%  &=Q \;\text{diag}\left( R'_{1} \dots R'_{4} \right)\bm{\lambda}\notag\\
%  &=\bm{Q'}\bm{\lambda},
% \end{align}
% where
% \begin{equation}
%  \bm{\lambda} = \left[
% \begin{array}{cccc}
%  \bm{\lambda_{1}}^T& \bm{\lambda_{2}}^T&\bm{\lambda_{3}}^T &\bm{\lambda_{4}} ^T\\
% \end{array}
% \right]^{T}.
% \end{equation} 
Consequently, the desired thrust allocation can be obtained as follows:
\begin{equation}\label{equation:des_lambda}
 {}^{\lbrace{CoG}\rbrace}\bm{\Lambda}^{des} = Q^{\#} {}^{\lbrace{CoG}\rbrace}\bm{W},
\end{equation}
where $Q^{\#}$ is the pseudo-inverse matrix of $Q$. Therefore, using \label{equation:integrated_r}, desired $i-th$ thrust vector can be obtained as follows:
\begin{equation}\label{equation:des_lambda}
 {}^{\lbrace{Rotor_{i}}\rbrace}\bm{\Lambda}^{des} = {}^{\lbrace Rotor_{i} \rbrace}R_{\lbrace CoG \rbrace } {}^{\lbrace{CoG}\rbrace}\bm{\Lambda}^{des},
\end{equation}
Lastly, the final control outputs including target thrust $\Lambda^{out}_{i}$ and tilt angle $\beta_{i}$ for each rotor can be calculate as following:
\begin{align}\label{equation:tilt_force}
 \Lambda^{out}_{i} = \|\bm{\Lambda}_{i}\|,\notag\\
 \beta_{i} = \text{atan}(\lambda_{i,z}, \lambda_{i,x}).
\end{align}

%Planning
\section{Reconfiguration Motion Planning}\label{sec:planning}
\begin{figure}[!t]
 \begin{center}
   \includegraphics[width=\columnwidth]{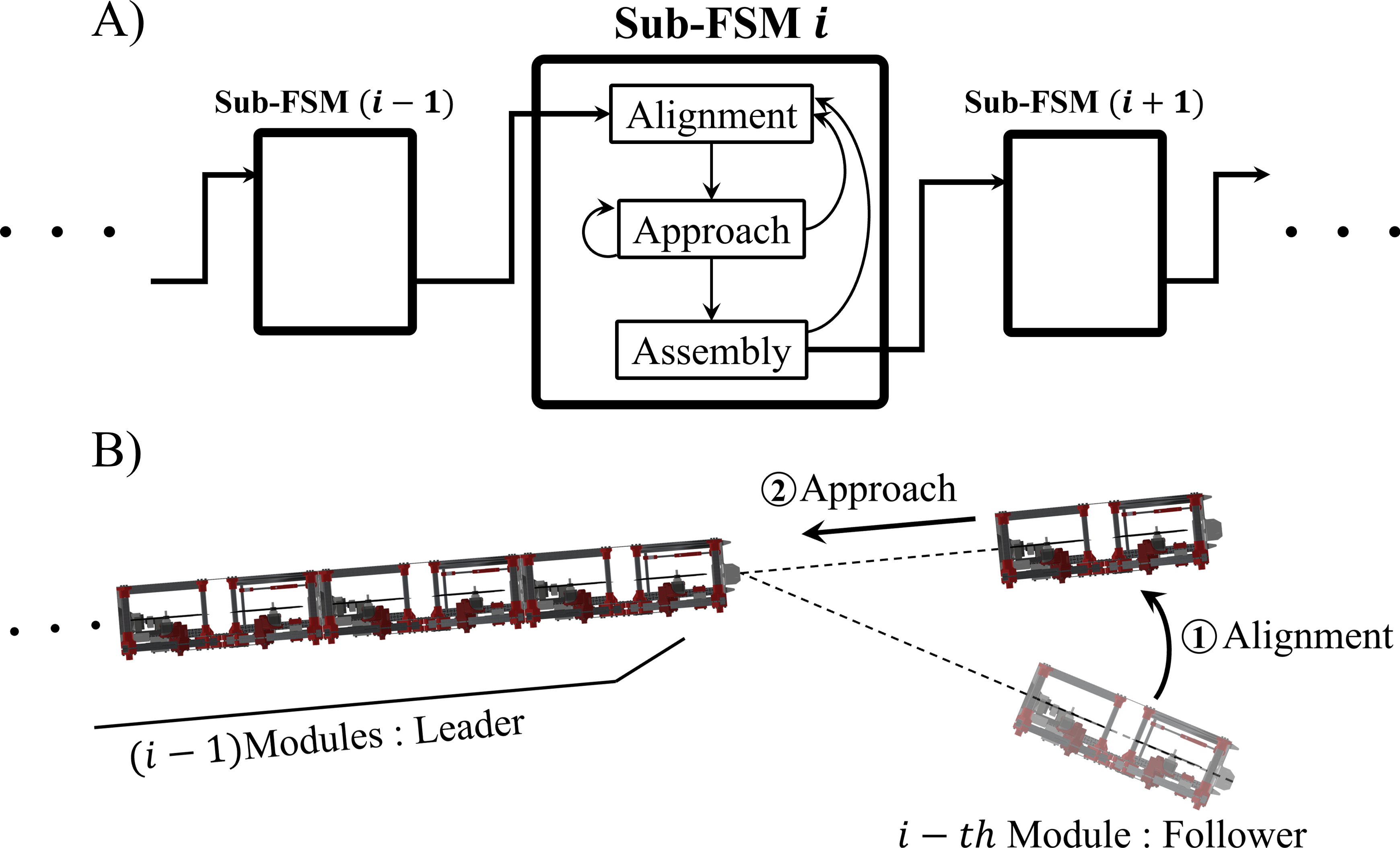}
 \end{center}
 \caption{The motion planning framework for aerial reconfiguration is depicted. In A), the comprehensive structure of the Finite State Machine (FSM) is delineated. Each block delineated with a standard line represents an individual state, whereas those with a bold line signify a Sub-FSM. These Sub-FSMs encompass states requisite for unitary reconfiguration motion. In B), the motion of robots throughout the processes within Sub-FSM $i$ is visually represented.}
   \label{fig:planning}
\end{figure}
In this section, we propose motion planning strategies aimed at achieving stable in-flight reconfiguration. As mentioned in our previous work \cite{sugihara2023}, pose control becomes unstable during aerial reconfiguration motion. Hence, Finite State Machine (FSM) style motion planning approach is appropriate since it can easily incorporate recovery actions.\par
As depicted in \figref{fig:planning} A), in this study, the planning process consists of multiple sub-FSMs to support reconfiguration motion with arbitrary number of modules. Each sub-FSM corresponds to a specific assembly motion, with the process transitioning to the next sub-FSM upon completion of the current one.\par

In a sub-FMS, the assembly motion is segmented into three states, as visualized in \figref{fig:planning} B). Initially, in the ``Approach'' state, the followr module is guided to a position and orientation that satisfies the following relationship with the leader structure:
\begin{align}\label{equation:align}
 &^{\lbrace GCoS \rbrace}\bm{r}_{\lbrace CoG_{i}\rbrace} =
  \left[
   \begin{array}{ccc}
    d^{st}&0&0
   \end{array}
 \right]^{T} , \notag\\
 &^{\lbrace GCoS \rbrace}R_{\lbrace CoG_{i} \rbrace} = \bm{0}
\end{align}
where $d^{st}$ is the desired distance between leader and follower.\par
If this relationship is achieved within an acceptable margin of error, the procedure progresses to the ``Approach'' state, where the follower moves towards the leader and endeavors to establish contact. During this motion, if the pose error exceeds an acceptable threshold, the reconfiguration process reverts to the ``Alignment'' state. Through this mechanism, it becomes possible to promptly rectify a hazardous pose situation between the two agents and resume the reconfiguration motion without interruption. \par
Finally, in the ``Assembly'' state, the leader and follower are fixed by activating the docking mechanism, followed by the updating of parameters such as the quantity of interconnected modules and their arrangement.\par
In the event of a disassembly motion, this can be achieved by transitioning these states in reverse order.

%multi
\section{Multi-Connected Flight}\label{sec:multi}
In this section, we introduce the methods to achieve stable multi-connected flight and aerial reconfiguration. Firstly, we present the method of target state conversion, followed by the framework for contact wrench compensation.
\subsection{Target State Conversion}
During multi-connected flight, the target position and attitude of the geometric center of the entire structure (GCoS) is transmitted from upstream. Hence, for the initial step, the target state of the GCoS should be transformed into the CoG of each module. This process corresponds to ``Target State Conversion'' block depicted in \figref{fig:framework}.\par
\begin{figure}[!t]
 \begin{center}
   \includegraphics[width=\columnwidth]{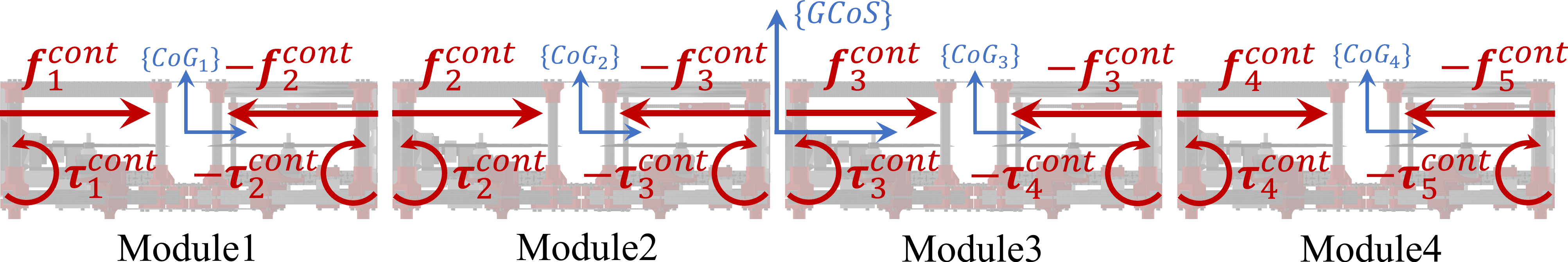}
 \end{center}
 \caption{Dynamics and coordinates model during four-connected flight.}
   \label{fig:dynamics}
\end{figure}

To begin, the position of the GCoS is defined as the average of the positions of CoGs of connected modules, as follows:
\begin{equation} \label{equation:gcos_r}
 ^{\lbrace W \rbrace}\bm{r}_{\lbrace GCoS \rbrace} = \frac{1}{n}\sum^{n}_{i = 1} {}^{\lbrace W \rbrace}\bm{r}_{\lbrace CoG \rbrace ,i},
\end{equation}
where $n$ represents the number of connected modules.
\figref{fig:dynamics} visualizes the relationship between $\lbrace CoG_{i} \rbrace $ in the case of four modules.
Subsequently, the conversion is derived using a homogeneous transformation matrix as follows:
\begin{align}\label{equation:target_conv}
& \left[
\begin{array}{c}
 ^{\lbrace W \rbrace}\bm{r}^{des}_{\lbrace CoG \rbrace , i}\\
 1\\
\end{array}
\right]
  = H_{i}
\left[
\begin{array}{c}
 ^{\lbrace W \rbrace}\bm{r}^{des}_{\lbrace GCoS \rbrace}\\
 1\\
\end{array}
\right], \\
&H_{i} = \left( \begin{array}{cc}
  {}^{\lbrace W\rbrace}\bm{R}^{des}_{\lbrace GCoS \rbrace}&\bm{d}^{3\times 1}_{i} \\
                                                               \bm{0}^{1\times 3}&1\\
                                                              \end{array}
\right), \notag \\
&\bm{d}_{i} ={}^{\lbrace W \rbrace}\bm{r}_{\lbrace CoG \rbrace ,i}- ^{\lbrace W \rbrace}\bm{r}_{\lbrace GCoS \rbrace}, \notag
\end{align}
where, ${}^{\lbrace W\rbrace}\bm{R}^{des}_{\lbrace GCoS \rbrace}$ represents the rotation matrix equivalent to the target Euler angle of the entire structure.\par

\subsection{Contact Wrench Compensation}\label{sec:wrench_comp}
The conversion process outlined in \eqref{equation:target_conv} theoretically generates target positions and orientations for each module that do not interfere with each other. However, in practice, unexpected contact wrenches accumulate between adjacent modules due to deviations in target poses among modules. This deviation arises from slight offsets in the sensors and subtle differences in aircraft characteristics between modules. The contact wrenches not only affect the stability of normal flight but also significantly destabilize the assembly and disassembly motion by abruptly appearing or disappearing during the process. Hence, we require a framework to dynamically and explicitly control this contact wrench as illustrated in B) block in \figref{fig:framework}. Following this, we elaborate on the design of the contact wrench filter and outline the method for calculating the feed-forward term.
\subsubsection{Dynamics in Multi-connected State}\label{sssec:1}
To simplify the dynamics model, we assume that the connection between modules is sufficiently strong and that no relative motion is generated between them. Therefore, according to the action-reaction law, the external wrench $\bm{W}^{ext}_{i}$ applied to the $i-th$ module can be expressed as follows:
\begin{equation}\label{equation:external_wrench}
\bm{W}^{ext}_{i} = \bm{W}_{i}^{cont}- \bm{W}_{i+1}^{cont} + \bm{W}^{dist},
\end{equation}
where $\bm{W}_{i}^{cont} = \left[ {}^{t}\bm{f}^{cont}_{i},\; {}^{t}\bm{\tau}^{cont}_{i} \right]^{T}$ is the contact wrench applied to the $i-th$ module by the $\left(i-1\right)-th$ module as illustrated in \figref{fig:dynamics}, and $\bm{W}^{dist}$ represents the disturbance wrench like aerodynamic effects. Note that in this study, we assume the disturbance wrench affects all modules equally.

\subsubsection{External Wrench Estimation}\label{sssec:2}
Generally, two methods are used to estimate external wrenches like $\bm{W}^{ext}$ in \eqref{equation:external_wrench}: momentum-based, utilizing velocity data, and acceleration-based. Due to substantial acceleration noise in the IMU sensors on Beatle, we employ the full momentum-based observer, as also utilized in \cite{Ryll2017}. By rewriting \eqref{equation:force} and \eqref{equation:torque} to incorporate external wrenches, we derive the following relationship:
\begin{align}\label{equation:euler_with_wext}
 \dot{\bm{p}} &= M\dot{\bm{v}} = J^{t}\bm{W} + \bm{W}^{ext} - N, \\
M &= \left(\begin{array}{cc}
 mE_{3\times 3}& 0_{3\times 3}\\
     0_{3\times 3}& I \\
    \end{array}\right),
J = \left(\begin{array}{cc}
 R^{T}& 0_{3\times 3}\\
     0_{3\times 3}&  E_{3\times 3}\\
    \end{array}\right), \notag\\
N &= \left(\begin{array}{c}
 m\bm{g}\\
      \bm{\omega}\times\;\bm{I}\;\bm{\omega}\\
    \end{array}\right)\notag,
\end{align}
where $\bm{p}$ and $\bm{v}$ represent the generalized momentum and velocity, respectively, while $\bm{W}$ and $\bm{W}^{ext}$ denote the wrench generated from rotor thrust and the external wrench, respectively. Subscripts and superscripts consistent with \eqref{equation:force} and \eqref{equation:torque} are omitted. According to \cite{luca2005}, the estimated external wrench can be expressed as the following residual term:

\begin{equation}\label{equation:wrench_est}
\hat{\bm{W}}^{ext} = K_{I}\left[ \bm{p}\left(t\right) - \bm{p}\left(t_{0}\right) - \int_{t_{0}}^{t}\left(J^{T} \bm{W} + \hat{ \bm{W}}^{ext} - N\right)dt  \right],
\end{equation}
where, $t$ and $t_{0}$ represent the current and initial time instances respectively, and $K_{I} \in R^{6\times 6}$ is a positive-definite gain matrix. Taking the time derivative of \eqref{equation:wrench_est}, we obtain the following dynamics for the residual vector:
\begin{equation}\label{equation:lpf}
\dot{\hat{\bm{W}}}^{ext} + K_{I}\hat{\bm{W}}^{ext} = K_{I}\bm{W}^{ext}.
\end{equation}
\eqref{equation:lpf} reveals a first-order low-pass filter that converges to $\bm{W}^{ext}$ as $t \to \infty$ approaches infinity, with the convergence speed dependent on the matrix $K_{I}$.\par
Note that this method only requires the estimation of the generalized velocity $\bm{v}$, which is less susceptible to sensor offsets than pose estimation.
\subsubsection{Contact Wrench Filtering}\label{sssec:3}
The quantity required for $i-th$ module to compensate for the contact wrench is $\bm{W}^{cont}_{i}$. However, $\hat{\bm{W}}^{ext}_{i}$ obtained in \eqref{equation:wrench_est} includes not only $\bm{W}^{cont}_{i}$ but also $\bm{W}^{cont}_{i+1}$ and $\bm{W}^{dist}$ as implied by \eqref{equation:euler_with_wext}. Therefore, a filter to remove these values from $\hat{\bm{W}}^{ext}_{i}$ is necessary. To decompose $\hat{\bm{W}}^{ext}_{i}$ into each element, the proposed filter utilizes the values of $\hat{\bm{W}}^{ext}_{k}$ from all other connected modules. According to \eqref{equation:euler_with_wext}, $\bm{W}^{dist}$ can be derived by summing up these estimated external wrench values as follows:
\begin{equation}
 \bm{W}^{dist} = \frac{1}{n}\sum^{n}_{k=1}\hat{\bm{W}}^{ext}_{k}.
\end{equation}
Note that each contact wrench cancels out, leaving only the $\bm{W}^{dist}$. Now, as $1st$ module is positioned at the left end, we have $\bm{W}^{cont}_{1} = 0$. Therefore, from \eqref{equation:euler_with_wext}, $\bm{W}^{cont}_{2}$ is obtained as follows.
\begin{equation}
 \bm{W}^{cont}_{2} = \hat{\bm{W}}^{ext}_{1} - \bm{W}^{dist}
\end{equation}
In a same manner, $\bm{W}^{cont}_{3},\;\bm{W}^{cont}_{4},\; \dots\bm{W}^{cont}_{n}$ are derived sequentially.\par
Note that this process necessitate communication among modules, hence the control frequency of the feedforward term relies on the bandwidth of this communication. However, it is crucial to emphasize that this process operates independently of normal flight control, thus any time delay does not adversely affect the frequency of flight control execution. Moreover, since we assume that the contact wrench is approximately steady in the case of BEATLE, any incurred time delay is not considered significant in ensuring stability.

\subsubsection{Feedforward Term for Wrench Compensation}\label{sssec:4}
To control the contact wrench, an appropriate feedforward term $\bm{W}^{ff} = \left[^{t}\bm{f}^{ff}\;\;^{t}\bm{\tau}^{ff}\right]^{T}$ is incorporated into the fundamental pose control output of each module as shown in \eqref{equation:des_f} and \eqref{equation:des_t}.\par
One important consideration is that this feedforward term has a cascading effect on other aircraft as well. First of all, the relationship between the $\bm{W}^{cont}_{i}$ at discrete times $t$ and $t+1$ is expressed by the following recurrence relation:
\begin{equation}\label{equation:recurrence}
 \bm{W}^{cont}_{i}(t+1) = \bm{W}^{cont}_{i}(t) - \bm{W}^{ff}_{i}(t+1) + \bm{W}^{ff}_{i-1} (t+1),
\end{equation}
where $\bm{W}^{ff}_{i}(t)$ is a feedforward wrench add to $i-th$ module at time $t$. Note that \eqref{equation:recurrence} includes $ \bm{W}^{ff}_{i-1}$ which is the feedforward term added to $(i-1)-th$ module. Here, \eqref{equation:recurrence} is subject to the following constraints:
\begin{align}
 \bm{W}^{cont}_{i}(t+1) &=  ^{des}\bm{W}^{cont}_{i}(t+1),\label{equation:c1}\\
 ^{des}\bm{W}^{cont}_{1}(t+1) &= \bm{W}^{cont}_{1}(t) = \bm{W}^{ff}_{0}(t+1) = \bm{0}, \label{equation:c2}
\end{align}
where $^{des}\bm{W}^{cont}_{i}(t+1)$ is the desired contact wrench occurring between module $(i-1)$ and module $i$, which is set to $\bm{0}$ in this study.\par
Under the conditions of \eqref{equation:c1} and \eqref{equation:c2}, solving the recurrence equation \eqref{equation:recurrence} yields the target feedforward term at time $t+1$ as follows:
\begin{equation}
 % \bm{W}^{ff}_{1}(t+1) =& \bm{0}, \notag \\ 
%  \bm{W}^{ff}_{2}(t+1) =& \bm{W}^{cont}_{2}(t) - ^{des}\bm{W}^{cont}_{2}(t+1), \notag \\ 
%  \bm{W}^{ff}_{3}(t+1) =& \sum^{3}_{k=1}\left(\bm{W}^{cont}_{k}(t) - ^{des}\bm{W}^{cont}_{k}(t+1) \right) , \notag \\
% &\vdots \notag \\
\bm{W}^{ff}_{i}(t+1) = \sum^{i}_{k=1}\lbrace \bm{W}^{cont}_{k}(t) - ^{des}\bm{W}^{cont}_{k}(t+1) \rbrace .
\end{equation}

 % This feedforward term $^{des}\bm{W}^{ff}_{i} = \left[{}^{t}\bm{f}^{ff}\;{}^{t}\bm{\tau}^{ff} \right]^{T}$ is computed for each module according to the following expression:
% \begin{equation}
%  ^{des}\bm{W}^{ff}_{i} = -^{des}\bm{W}^{cont}_{i} + \bm{W}^{cont}_{i}
% \end{equation}
% where $^{des}\bm{W}^{cont}_{i}$ represents the target contact wrench applied between the $i-th$ and $\left(i-th\right)$ modules, which is set to $0$ in this study.
% It is important to note that $^{des}\bm{W}^{ff}_{i}$ affects $\bm{W}^{cont}_{i+1}$ in a chain. That is, in discrete time, adding $^{des}\bm{W}^{ff}_{2}\left(t\right)$ to the $2nd$ module, for instance, results in $\bm{W}^{cont}_{3}\left(t + 1\right) = \bm{W}^{cont}_{3}\left(t\right) + ^{des}\bm{W}^{ff}_{2}\left(t\right)$. Considering this cascading effect, the final feed-forward term to be given to each module is as follows:
% \begin{equation}
%  ^{des}\bm{W'}^{ff}_{i} = \sum^{i-1}_{k=1} {}^{des}\bm{W}'^{ff}_{k} + \left(-^{des}\bm{W}^{cont}_{i} + \bm{W}^{cont}_{i}\right).
% \end{equation}

%simulation
\section{Simulation Studies} \label{sec:simulation}
In this study, we conduct simulation experiments that are challenging to carry out in the real world due to safety and resource constraints.
\subsection{Simulation Setup}
% \begin{table}[!t]
%   \caption{parameter for simulation noise}
%   \centering
%  \begin{tabular}{ccc}
%   \hline
%   State & Mean & Standard deviation\\
%   \hline
%   Position & $0.02$ & $0.01$\\
%   Rotation & $0.1$ & $0.01$\\
% \hline
%  \end{tabular}
%  \label{table:noise_param}
% \end{table}
Regarding the simulation conditions, we utilize Gazebo, ensuring that each physical parameters matches those of the real machine. Furthermore, we introduce thfe Gaussian noise to the pose information obtained from Gazebo to replicate sensor noise in the real machine. The mean and standard deviation of the noise are $0.05$ and $0.01$, respectively.
\subsection{Validation of Contact Wrench Compensation}
\begin{figure}[!t]
 \begin{center}
   \includegraphics[width=\columnwidth]{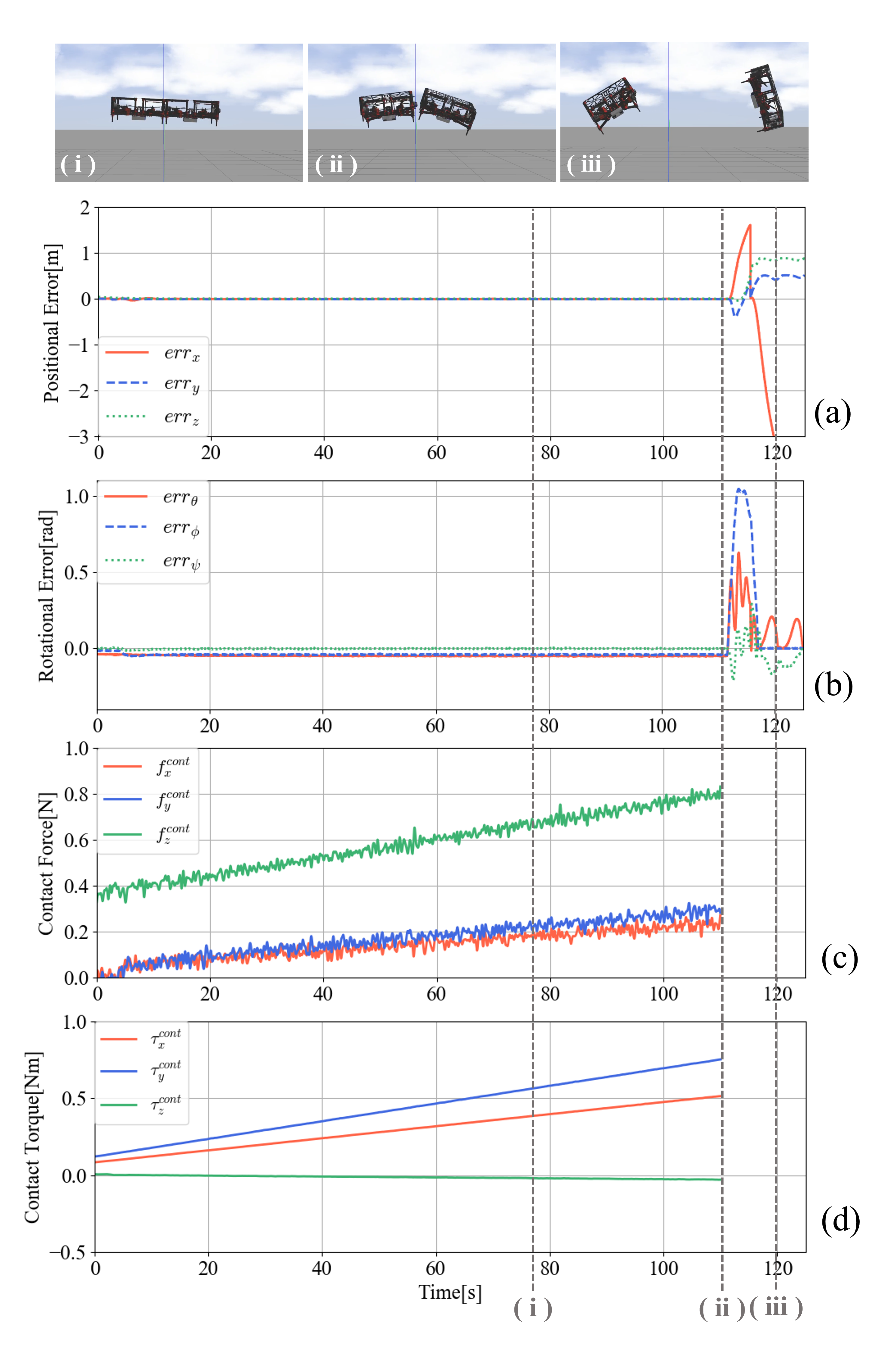}
 \end{center}
 \caption{The Results of multi-connected hovering and in-flight separation simulation without the proposed wrench compensation method: (a) and (b) show the positional and rotational errors of the left module. Note that only the errors in the left module are shown for simplicity; (c) and (d) represent the contact force and torque occurring between two modules. (i) through (iii) in the graph correspond to (i) through (iii) in the snapshots.}
   \label{fig:sim_disassembly_wo_cwc}
\end{figure}
% \begin{figure}[!t]
%  \begin{center}
%    \includegraphics[width=\columnwidth]{disassembly_sim_w_cwc.png}
%  \end{center}
%  \caption{The Results of multi-connected hovering and in-flight separation simulation with the wrench compensation method. The caption follows the same rule as in \figref{fig:sim_disassembly_wo_cwc}.}
%    \label{fig:sim_disassembly_w_cwc}
% \end{figure}
Firstly, we validate our hypothesis regarding contact wrench accumulation and the effectiveness of the proposed method in eliminating this contact wrench. Note that, similar to real world, the self-position estimation of each module is slightly deviated due to sensor noise in the simulation. \figref{fig:sim_disassembly_wo_cwc} displays the outcomes of multi-connected flight and disassembly motion without the proposed contact wrench compensation. According to \figref{fig:sim_disassembly_wo_cwc} (c) and (d), it is evident that the contact wrench gradually accumulates at the coupling between two modules. As \figref{fig:sim_disassembly_wo_cwc} (a) and (b) reveal, when the coupling is released, the accumulated wrench is released, causing the divergence of pose control of each module.\par
% On the other hand, \figref{fig:sim_disassembly_w_cwc} illustrates the result when the proposed method is applied. From \figref{fig:sim_disassembly_w_cwc} (c) and (d), it can be observed that the generation of contact wrench is suppressed. Additionally, from \figref{fig:sim_disassembly_w_cwc} (a) and (b), it is evident that each module can execute disassembly motion stably.

\subsection{Scaling Feasibility Verification}
\begin{figure}[!t]
 \begin{center}
   \includegraphics[width=\columnwidth]{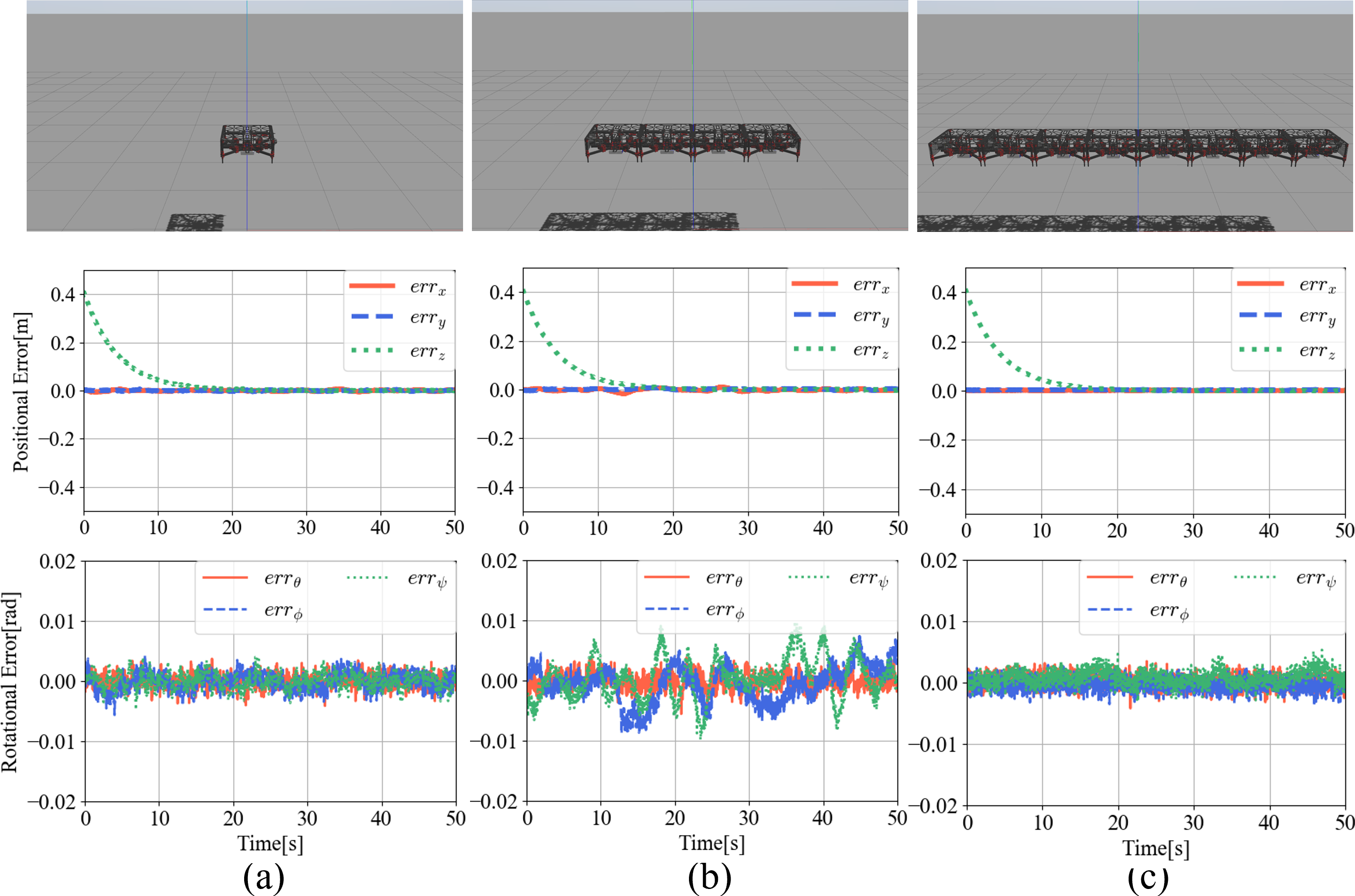}
 \end{center}
 \caption{The results of takeoff and hovering simulation with different numbers of connected modules: (a), (b), and (c) show the results for one, four, and eight moudules, respectively. Regarding the positional and rotational errors, the values for the module closest to the center of the structure are shown as representative values.}
   \label{fig:sim_scaling}
\end{figure}

\begin{table}[b]
  \caption{RMSE during hovering in each numer of connected modules}
  \centering
 \begin{tabular}{c cccccc}
  \hline
  \multirow{2}{*}{n} & \multicolumn{6}{c}{RMSE}\\
  \cline{2-7}
   & $e_{x}$[m] & $e_{y}$[m] & $e_{z}$[m]  & $e_{\theta}$[rad]  & $e_{\phi}$[rad]  & $e_{\psi}$[rad]\\ \hline
  1&2.0e-3&1.9e-3&3.8e-3&1.2e-3 &1.5e-3 &1.3e-3\\ 
  4&4.0e-3&3.1e-3&4.0e-3&1.4e-3 &3.1e-3 &3.5e-3\\ 
  8&1.4e-3&3.9e-3&3.0e-3&1.0e-3 &1.1e-3 &1.5e-3\\
  \hline
 \end{tabular}
 \label{table:scaling_sim}
\end{table}
Next, we verify the scalability of our system by comparing the stability of flight with different numbers of connected modules. In this simulation, takeoff and hovering motions with proposed contact wrench compensation method are observed. The results are shown in \figref{fig:sim_scaling} and \tabref{table:scaling_sim}. From the results in \tabref{table:scaling_sim}, it is evident that increasing the number of connected aircraft does not affect the accuracy of pose control. This implies that the proposed system is scalable to an arbitrary number of modules.

%experiment
\section{Experiment}\label{sec:experiment}
\subsection{Implementation}
\begin{figure}[!t]
 \begin{center}
   \includegraphics[width=\columnwidth]{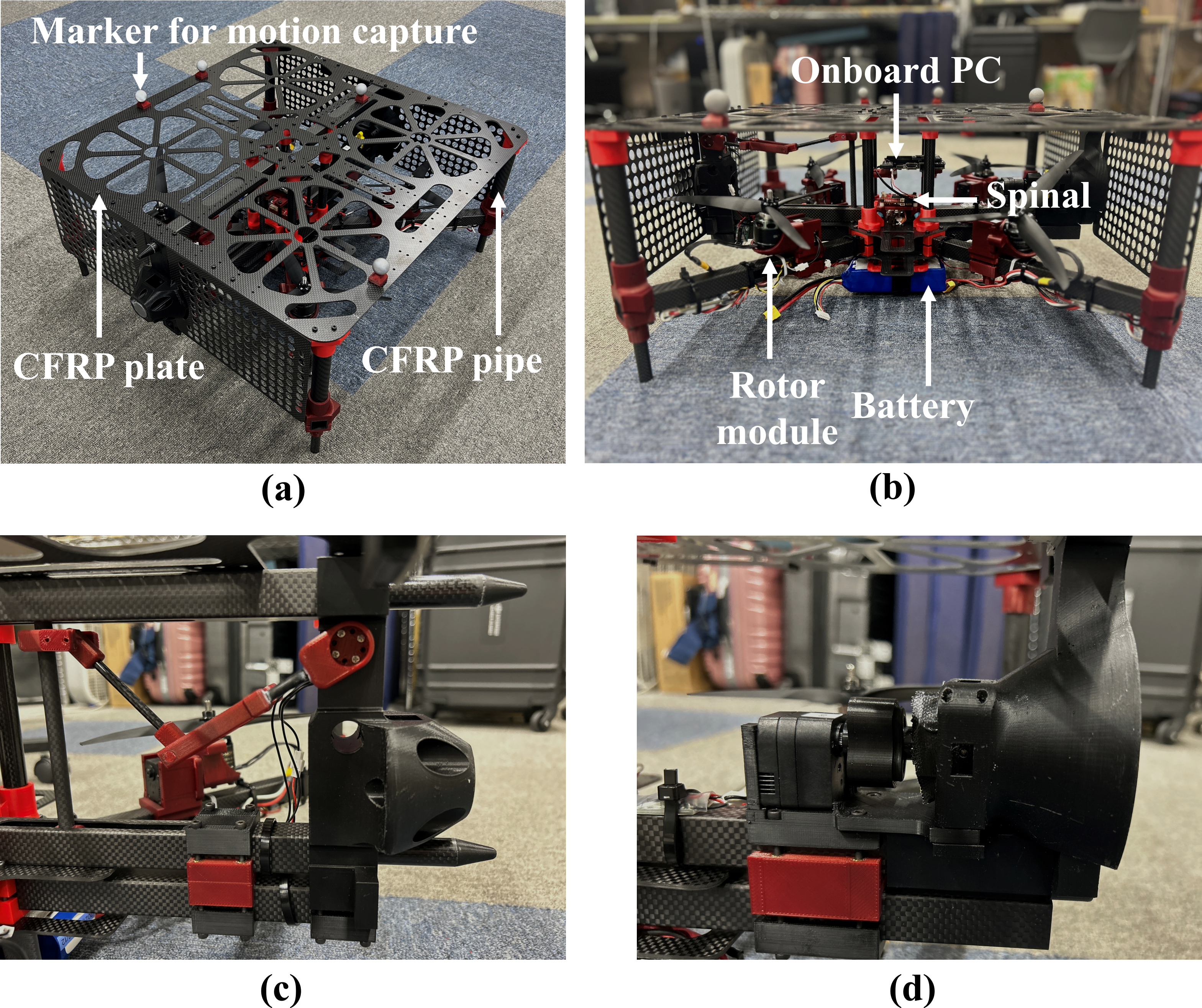}
 \end{center}
 \caption{The snapshot of the prototype of BEATLE module. (a): Exterior. (b): Detailed components. (c): Male-side docking mechanism. (d): Female-side docking mechanism.}
   \label{fig:implementation}
\end{figure}
\subsubsection{Hardware Configuration}
Based on the design proposed in \secref{sec:design}, we implemente the prototype of BEATLE as depicted in \figref{fig:implementation}. \par
The skeleton and housing primarily consist of CFRP, while other components are constructed from PLA. The CFRP plate has a thickness of \SI{2}{mm}, and the CFRP cylinder has a thickness of \SI{1}{mm}.\par

The rotor module consists of a 9-inch propeller (GEMFAN 9045-3), a brushless motor (T-Motor AT2814), and a servo motor (KRS3304R2). Each rotor is driven by an ESC (T-Motor AIR 40A). With this configuration, each rotor generates thrust ranging from approximately \SI{1}{N} to \SI{20}{N} at a voltage of \SI{25.2}{V} (\SI{6}{S}).
\par

Regarding processors, our original flight control unit, named ``Spinal,'' is installed, equipped with an onboard IMU and an STM32 MUC. The ``Fundamental Control Flow'' in \figref{fig:framework} runs at a frequency of \SI{500}{Hz} on the Spinal. Additionally, as a high-level processor, an onboard PC (VIM 4) is equipped, where ``Trajectory Conversion'' and ``Contact Wrench Compensation'' in \figref{fig:framework} operate at \SI{40}{Hz}.
\par

The primary characteristics of the developed BEATLE prototype are presented in \tabref{table:hard_prop}.

\subsubsection{External Components}
In the real-machine experiment, we utilize a motion capture system for position estimation and a laptop to transmit upstream commands. Communication between these external components and the robot occurs via a Wi-Fi network.
\begin{table}[!t]
  \caption{main parameters of robot}
  \centering
 \begin{tabular}{cc}
  \hline
  Parameter&Value \\
  \hline
  Mass of main body& \SI{4.2}{kg}\\
  Size of main body& \SI{0.52}{m} $\times$\SI{0.52}{m}\\
  Battery&\SI{6000}{mAh} \\
  Continuous flight time&\SI{6}{min}\\
  \hline
 \end{tabular}
 \label{table:hard_prop}
\end{table}

\subsection{Trajectory Tracking}
\begin{table}[b]
  \renewcommand{\arraystretch}{1.3}
  \caption{RMSE during trajectory tracking in each numer of connected modules}
  \centering
 \begin{tabular}{cccc}
  \hline
  \multirow{2}{*}{n} & \multicolumn{3}{c}{RMSE}\\
  \cline{2-4}
   & $e_{\theta}$[rad] & $e_{\phi}$[rad] & $e_{\psi}$[rad]\\ \hline
  1&1.5e-2&1.9e-2&4.3e-2\\ 
  2&1.7e-2&2.8e-2&4.8e-2\\ 
  3&2.1e-2&4.8e-2&4.3e-2\\
  \hline
 \end{tabular}
 \label{tab:lemniscate}
\end{table}
  \renewcommand{\arraystretch}{1.0}
\begin{figure}[!t]
 \begin{center}
   \includegraphics[width=\columnwidth]{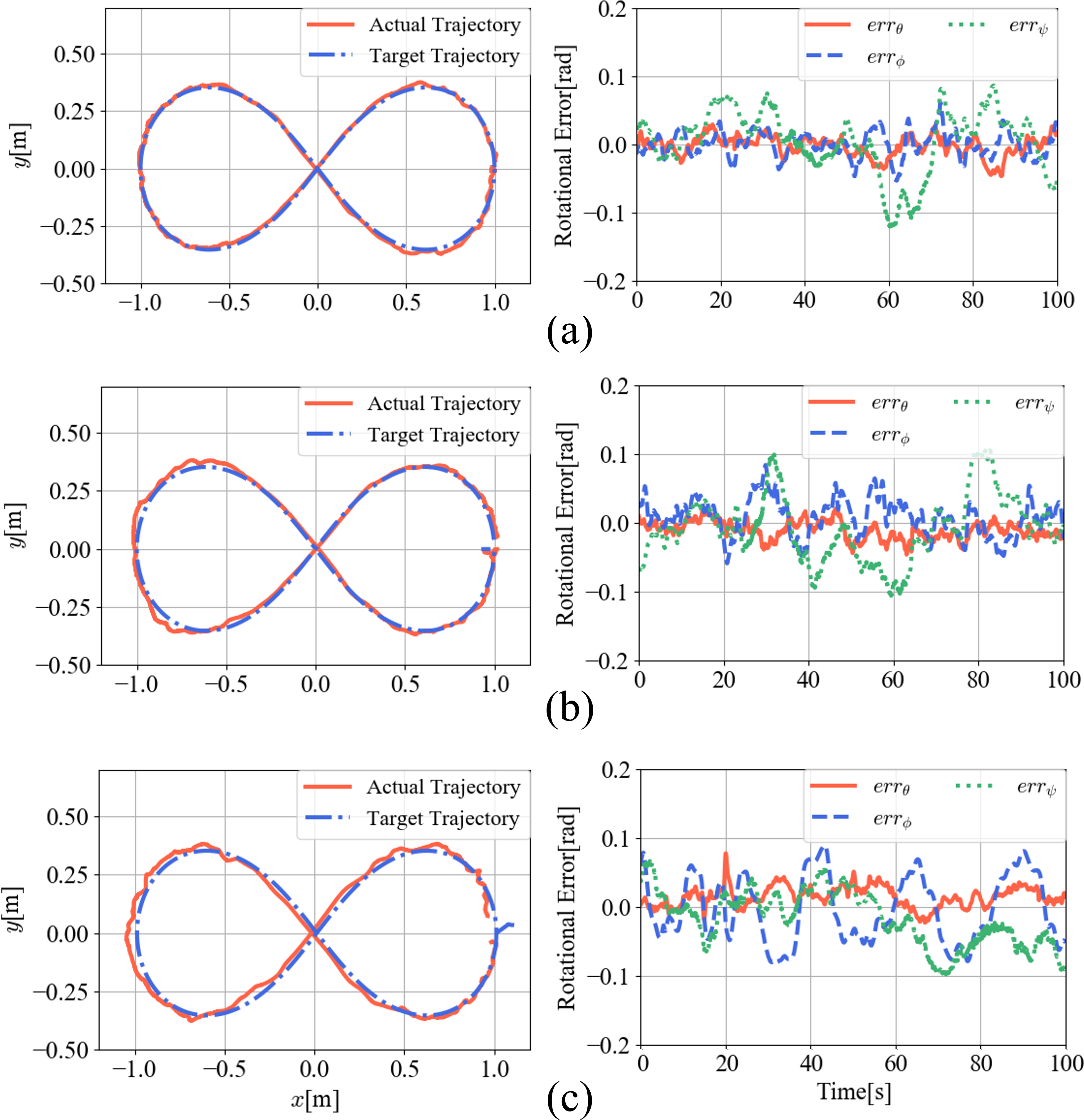}
 \end{center}
 \caption{The results of trajectory tracking experiments with different numbers of connected modules. (a), (b), and (c) show the results for one, two, and three modules, respectively. Left: visualized trajectory tracking; right: in-flight attitude control error.}
   \label{fig:lemniscate}
\end{figure}
We begin by introducing the trajectory tracking experiments. In this experiment, various numbers of connected modules track a lemniscate trajectory defined as $r^2 = \sqrt{\cos\left(2\theta\right)}$. Here, $\theta$ transitions $0 \to 2\pi$ over \SI{100}{s} at a constant rate. Consequently, the target translational acceleration changes nonlinearly.\par

In this experiment, we utilize 1 to 3 modules. The results for each number of connected modules are depicted in \figref{fig:lemniscate} and \tabref{tab:lemniscate}. \par
From these results, it can be observed that even with an increase in the number of connected modules, the modules can achieve various accelerations while maintaining flight stability. Although the rotational error slightly increases as the number of modules increases, this is primarily due to the increase in aerodynamic interference caused by the surrounding walls and is considered specific to this experimental environment.

\subsection{In-flight assembly and disassembly}
\begin{table}[b]
  \renewcommand{\arraystretch}{1.3}
  \caption{RMSE during reconfiguration motion}
  \centering
 \begin{tabular}{cccccc}
  \hline
   $e_{x}$[m] & $e_{y}[m]$ & $e_{z}[m]$ &$e_{\theta}[rad]$ & $e_{\phi}[rad]$ & $e_{\psi}[rad]$\\ \hline
  2.3e-2&2.2e-2&2.5e-2&1.6e-2&5.2e-2&4.8e-2\\ 
  \hline
 \end{tabular}
 \label{tab:reconfig}
\end{table}
  \renewcommand{\arraystretch}{1.0}
Next, we demonstrate an in-flight reconfiguration experiment. In this experiment, three modules merge sequentially, and subsequently, they separate sequentially.\par
\begin{figure}[!t]
 \begin{center}
   \includegraphics[width=\columnwidth]{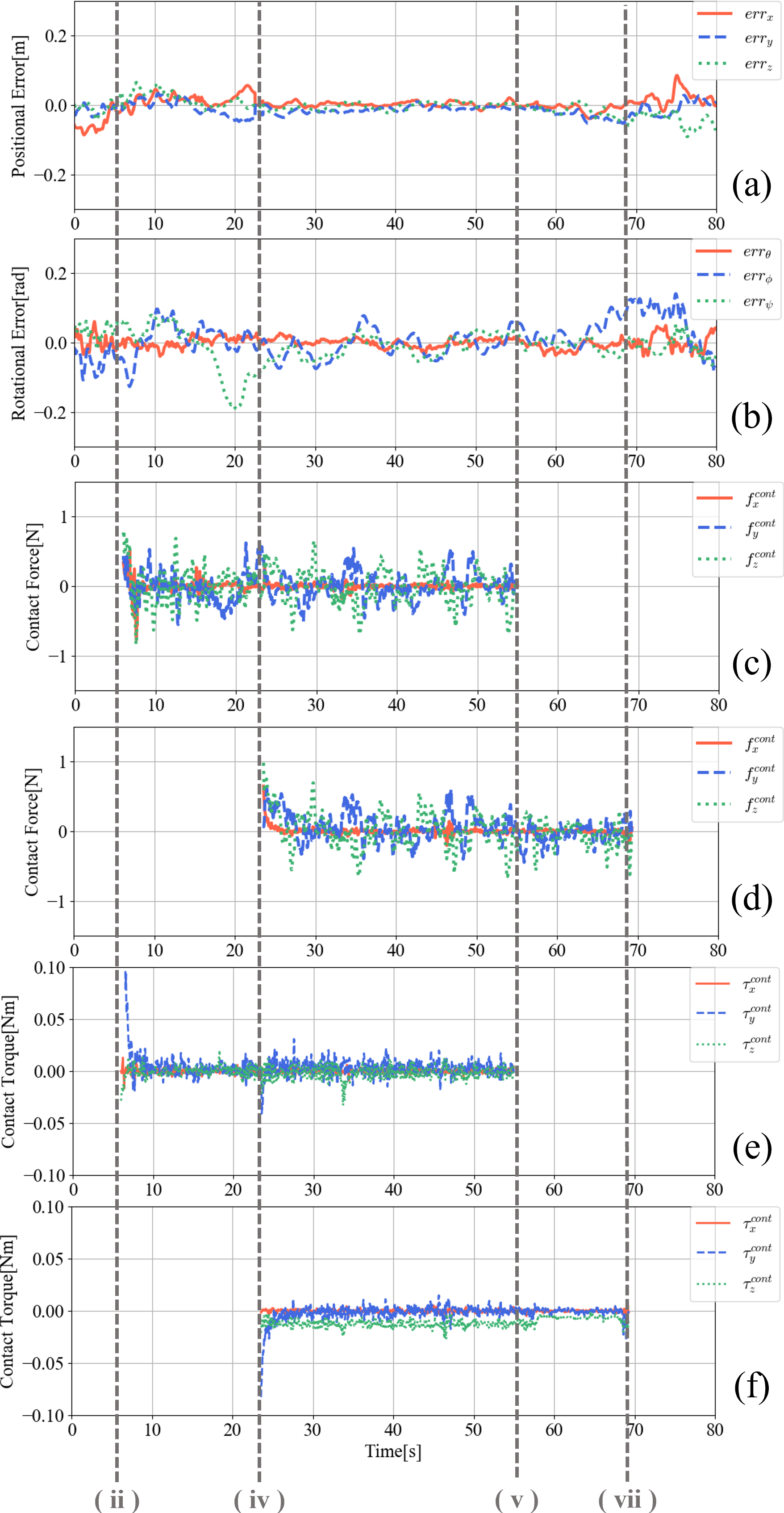}
 \end{center}
 \caption{The Results of the in-flight reconfiguration experiment: (a) and (b) show the positional and rotational errors. Note that the values for the module closest to the center of the structure are shown as representative values; (c) and (d) represent the contact force and torque occurring between module 1 and module 2, while (e) and (f) represent those between module 2 and module 3. (ii) through (vii) in the graph correspond to (ii) through (vii) in the snapshots.
 }
   \label{fig:reconfig_graph}
\end{figure}
\begin{figure}[!t]
 \begin{center}
   \includegraphics[width=\columnwidth]{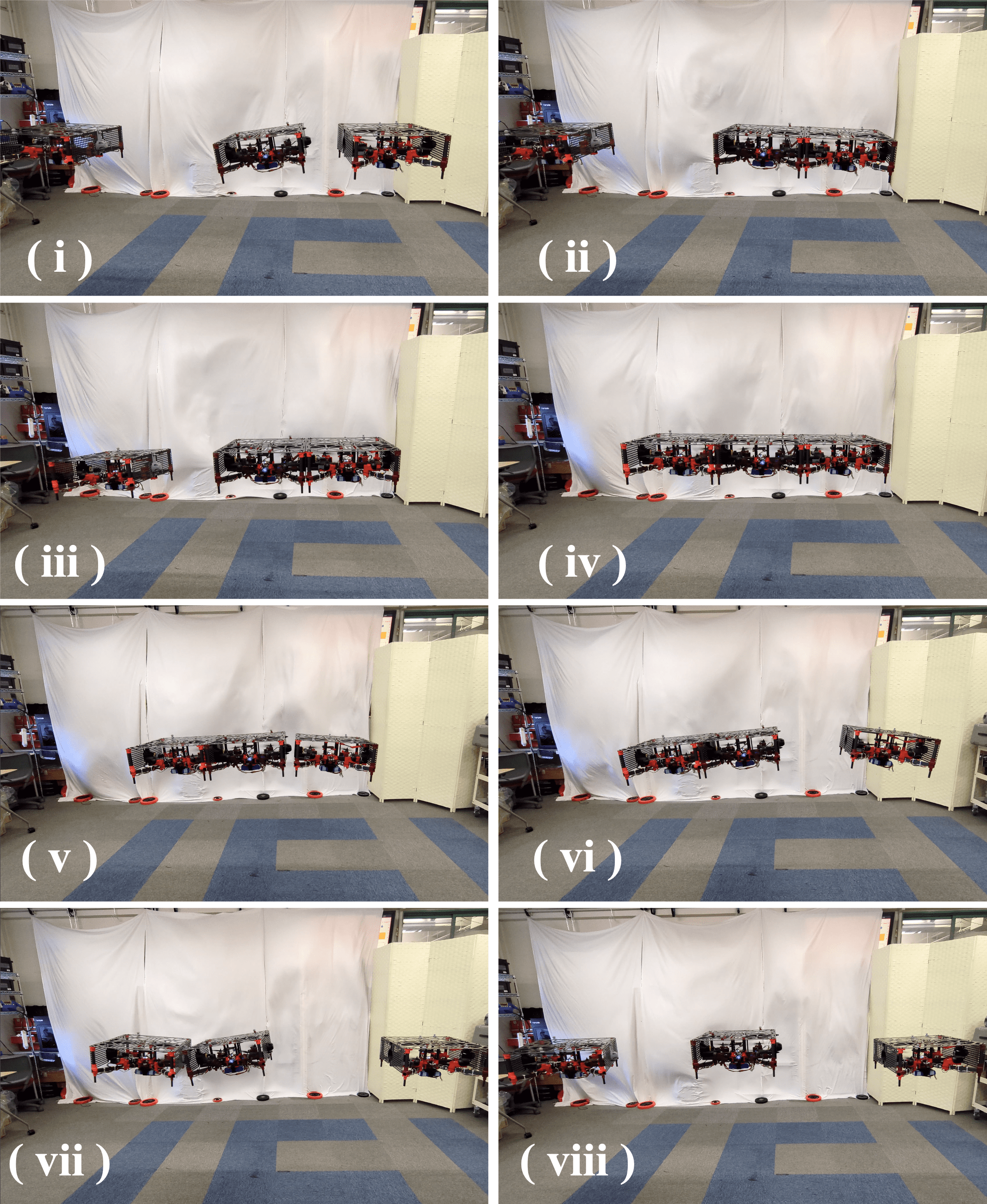}
 \end{center}
 \caption{The snapshots of the in-flight reconfiguration experiment. From left to right, the modules are designated as module1, module2, and module3. (i)-(iv) represent assembly motion, while (v)-(viii) depict disassembly motion.}
   \label{fig:reconfig_snap}
\end{figure}

The result of the experiment is depicted in \figref{fig:reconfig_graph}, \figref{fig:reconfig_snap}, and \tabref{tab:reconfig}. According to \figref{fig:reconfig_graph} (c)-(f), it can be observed that the contact wrench between modules instantaneously occurs when a module docks, but this wrench is instantly eliminated, allowing the merged modules to continue stable flight. This result indicates that the proposed wrench compensation method functioned properly. \figref{fig:reconfig_graph} (a) and (b) also demonstrate that pose instability, similar to those observed in previous conventional A-MSRRs \cite{Saldana2018,moddessemble,sugihara2023}, does not occur during reconfiguration motions. \par
On the other hand, slight instabilities in position and orientation can be observed just before assembly (around \SI{20}{s}) and just after separation (around \SI{70}{s}). These instabilities are presumably caused by aerodynamic interference between modules that are very close together.\par

In this experiment, hovering was sustained for more than \SI{30}{s} after the three bodies merged. However, no accumulation of contact wrenches was observed and the position error and attitude error were both within \SI{0.025}{m} and \SI{0.048}{rad}, respectively.

%conclusion
\section{Conclusion}\label{sec:conclusion}
In this paper, we present the design and control framework for a Aerial Modular Self-Reconfigurable Robot (A-MSRR) named BEATLE, capable of connecting and separating in-flight. 
BEATLE is equipped with a coupling mechanism consisting of both magnetic and mechanical locking mechanisms, as well as a robust housing, enabling the construction of a highly rigid structure.
By proposing a thrust configuration with vectoring rotors that enables over-actuation and a control framework for multi-connected flight incorporating contact wrench controller, we have significantly enhanced the scalability and stability of conventional methods.
The feasibility and scalability of the proposed framework have been validated through simulation studies and experiments with implemented prototype robots, which encompass hovering, trajectory tracking, and in-flight self-reconfiguration.\par

Although we solely focus on A-MSRR with rigid and one-direction connections in this work, it is expected that the proposed method can be extended to A-MSRR systems with joint and multi-direction connections in future work. With the extended A-MSRR system, various structures such as bridges with complicated truss structures, multi-link arms connected to ceilings, and multi-legged robots capable of terrestrial movement can be realized.\par
\FloatBarrier

% \appendices
% \section{Proof of the First Zonklar Equation}
% Appendix one text goes here.

% % you can choose not to have a title for an appendix
% % if you want by leaving the argument blank
% \section{}
% Appendix two text goes here.

% % use section* for acknowledgment
% \section*{Acknowledgment}

% The authors would like to thank...

% Can use something like this to put references on a page
% by themselves when using endfloat and the captionsoff option.
% \ifCLASSOPTIONcaptionsoff
%   \newpage
% \fi

% trigger a \newpage just before the given reference
% number - used to balance the columns on the last page
% adjust value as needed - may need to be readjusted if
% the document is modified later
%\IEEEtriggeratref{8}
% The "triggered" command can be changed if desired:
%\IEEEtriggercmd{\enlargethispage{-5in}}

% references section

% can use a bibliography generated by BibTeX as a .bbl file
% BibTeX documentation can be easily obtained at:
% http://mirror.ctan.org/biblio/bibtex/contrib/doc/
% The IEEEtran BibTeX style support page is at:
\bibliographystyle{IEEEtran}
% http://www.michaelshell.org/tex/ieeetran/bibtex/
% argument is your BibTeX string definitions and bibliography database(s)
\bibliography{IEEEabrv,bib}
% \printbibliography
%
% <OR> manually copy in the resultant .bbl file
% set second argument of \begin to the number of references
% (used to reserve space for the reference number labels box)
% \begin{thebibliography}{1}

% \bibitem{IEEEhowto:kopka}
% H.~Kopka and P.~W. Daly, \emph{A Guide to \LaTeX}, 3rd~ed.\hskip 1em plus
%   0.5em minus 0.4em\relax Harlow, England: Addison-Wesley, 1999.

% \end{thebibliography}

% biography section
%
% If you have an EPS/PDF photo (graphicx package needed) extra braces are
% needed around the contents of the optional argument to biography to prevent
% the LaTeX parser from getting confused when it sees the complicated
% \includegraphics command within an optional argument. (You could create
% your own custom macro containing the \includegraphics command to make things
% simpler here.)
%\begin{IEEEbiography}[{\includegraphics[width=1in,height=1.25in,clip,keepaspectratio]{mshell}}]{Michael Shell}
% or if you just want to reserve a space for a photo:

% \begin{IEEEbiography}{Michael Shell}
% Biography text here.
% \end{IEEEbiography}

% if you will not have a photo at all:
% \begin{IEEEbiographynophoto}{John Doe}
% Biography text here.
% \end{IEEEbiographynophoto}

% \begin{IEEEbiographynophoto}{Jane Doe}
% Biography text here.
% \end{IEEEbiographynophoto}

\begin{IEEEbiography}[{\includegraphics[width=1in,height=1.25in,clip,keepaspectratio]{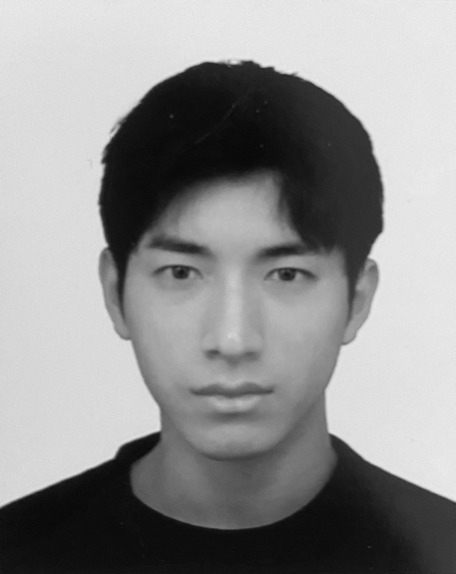}}]{Junichiro Sugihara}
 is a PhD student in the Department of Mechano-Informatics, School of Information Science and Technology, University of Tokyo. He received BE degree from the Department of Mechanical Engineering, University of Tokyo, in 2023. His research interests are mechanical design, modeling and control of aerial robots, and self-reconfigurable modular robot.
\end{IEEEbiography}

\begin{IEEEbiography}[{\includegraphics[width=1in,height=1.25in,clip,keepaspectratio]{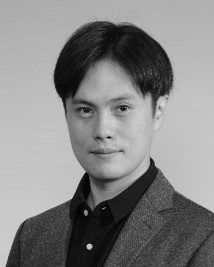}}]{Moju Zhao}
 is currently an Associate Professor at The University of Tokyo. He received Doctor Degree from the Department of Mechano-Informatics, The University of Tokyo, 2018. His research interests are mechanical design, modelling and control, motion planning, and vision based recognition in aerial robotics. His main achievement is the articulated aerial robots which have received several awards in conference and journal, including the Best Paper Award in IEEE ICRA 2018.
\end{IEEEbiography}

\begin{IEEEbiography}[{\includegraphics[width=1in,height=1.25in,clip,keepaspectratio]{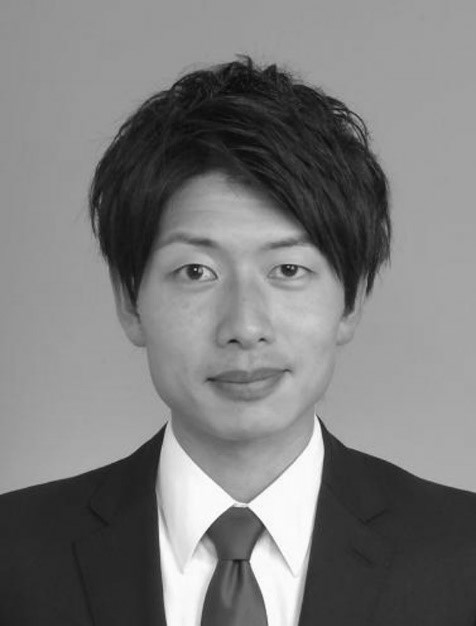}}]{Takuzumi Nishio}
 is an Assistant Professor in the Department of Mechano-Informatics, School
of Information Science and Technology, University of Tokyo. He received PhD degree from the Department of Interdisciplinary-Informatics, University of Tokyo, in 2022. His research interests are mechanical design, modeling and control of aerial robots, and vision-based recognition and motion planning of manipulators in field robotics.
\end{IEEEbiography}

\begin{IEEEbiography}[{\includegraphics[width=1in,height=1.25in,clip,keepaspectratio]{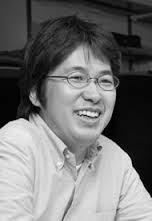}}]{Kei Okada}
 received BE in Computer Science from Kyoto University in 1997. He received MS and PhD in Information Engineering from The University of Tokyo in 1999 and 2002, respectively. From 2002 to 2006, he joined the Professional Programme for Strategic Software Project in The University Tokyo. He was appointed as a lecturer in the Creative Informatics at the University of Tokyo in 2006 and an associate professor in the Department of Mechano-Informatics in 2009. His research interests include humanoids robots, real-time 3D computer vision, and recognitionaction integrated system.
\end{IEEEbiography}

\begin{IEEEbiography}[{\includegraphics[width=1in,height=1.25in,clip,keepaspectratio]{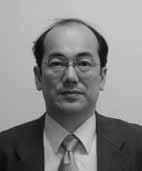}}]{Masayuki Inaba}
 is a professor in the Department of Mechano-Informatics, School of Information Science and Technology, University of Tokyo. He received PhD degree from the Graduate School of Information Engineering, University of Tokyo, in 1986. He is currently a professor in the Department of Mechano-Informatics, University of Tokyo, in 2000. His research interests include robotic systems and software architectures to advance robotics research.
 \end{IEEEbiography}
\end{document}